\documentclass[a4paper,fleqn]{cas-dc} 

\usepackage[numbers, authoryear,longnamesfirst]{natbib}
\usepackage{url}
\usepackage{multirow}
\usepackage{multicol}
\usepackage{booktabs}
\usepackage{color}
\usepackage{array}
\usepackage{amsmath}
\usepackage{cuted}
\usepackage{amssymb}
\usepackage{mathtools}
\usepackage{float}
\usepackage{stfloats}
\usepackage{xspace}
\usepackage{silence}
\usepackage{graphicx}
\usepackage{subfigure}
\usepackage[utf8]{inputenc}
\usepackage{dirtytalk} 
\usepackage{makecell}

\def\tsc#1{\csdef{#1}{\textsc{\lowercase{#1}}\xspace}}
\tsc{WGM}
\tsc{QE}
\tsc{EP}
\tsc{PMS}
\tsc{BEC}
\tsc{DE}

\begin{document}
\let\WriteBookmarks\relax
\def\floatpagepagefraction{1}
\def\textpagefraction{.001}

\shorttitle{DAM-Net}

\shortauthors{Tamer~Saleh et~al.}

\title [mode = title]{DAM-Net: Global Flood Detection from SAR Imagery Using Differential Attention Metric-Based Vision Transformers}                      



\author[1,3]{Tamer Saleh}
\fnmark[1]
\ead{tamersaleh@whu.edu.cn}

\author[4]{Xingxing Weng}
\fnmark[1]
\ead{xingxingw@whu.edu.cn}

\author[1]{Shimaa Holail}
\ead{shimaaholail@whu.edu.cn}

\author[4]{Chen Hao}
\ead{chenhao20@whu.edu.cn}

\author[1,2,4]{Gui-Song Xia}
\cormark[1]
\ead{guisong.xia@whu.edu.cn}

\affiliation[1]{organization={State Key Laboratory of Information Engineering in Surveying, Mapping and Remote Sensing (LIESMARS), Wuhan University},
    city={Wuhan},
    postcode={430079}, 
    country={China}}

\affiliation[2]{organization={National Engineering Research Center for Multi-media Software, School of Computer Science and Institute of Artificial Intelligence, Wuhan University},
    city={Wuhan},
    postcode={430072}, 
    country={China}}
    
\affiliation[3]{organization={Geomatics Engineering Department, Faculty of Engineering at Shoubra, Benha University},
    city={Cairo},
    postcode={1196}, 
    country={Egypt}}

\affiliation[4]{organization={School of Computer Science, Wuhan University},
    city={Wuhan},
    postcode={430072}, 
    country={China}}

\cortext[cor1]{Corresponding author}
\fntext[fn1]{have the same contribution to this work.}


\begin{abstract}
The detection of flooded areas using high-resolution synthetic aperture radar (SAR) imagery is a critical task with applications in crisis and disaster management, as well as environmental resource planning. However, the complex nature of SAR images presents a challenge that often leads to an overestimation of the flood extent. To address this issue, we propose a novel differential attention metric-based network (DAM-Net) in this study. The DAM-Net comprises two key components: a weight-sharing Siamese backbone to obtain multi-scale change features of multi-temporal images and tokens containing high-level semantic information of water-body changes, and a temporal differential fusion (TDF) module that integrates semantic tokens and change features to generate flood maps with reduced speckle noise. Specifically, the backbone is split into multiple stages. In each stage, we design three modules, namely, temporal-wise feature extraction (TWFE), cross-temporal change attention (CTCA), and temporal-aware change enhancement (TACE), to effectively extract the change features. In TACE of the last stage, we introduce a class token to record high-level semantic information of water-body changes via the attention mechanism. Another challenge faced by data-driven deep learning algorithms is the limited availability of flood detection datasets. To overcome this, we have created the \textit{S1GFloods} open-source dataset, a global-scale high-resolution Sentinel-1 SAR image pairs dataset covering 46 global flood events between 2015 and 2022. The experiments on the \textit{S1GFloods} dataset using the proposed DAM-Net showed top results compared to state-of-the-art methods in terms of \textit{overall accuracy}, \textit{F1}-score, and \textit{IoU}, which reached 97.8\%, 96.5\%, and 93.2\%, respectively. Our dataset and code will be available online at \url{https://github.com/Tamer-Saleh/S1GFlood-Detection}.
\end{abstract}

\maketitle

\section{Introduction}
\label{Introduction}

\begin{figure*}[!htb]
    \centering
    \subfigure[]{\includegraphics[width=0.150\textwidth]{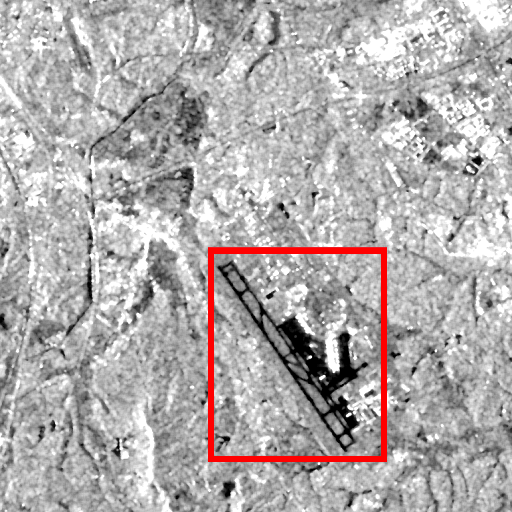}
    \label{fig: intro1}}
    \hfil
    \subfigure[]{\includegraphics[width=0.150\textwidth]{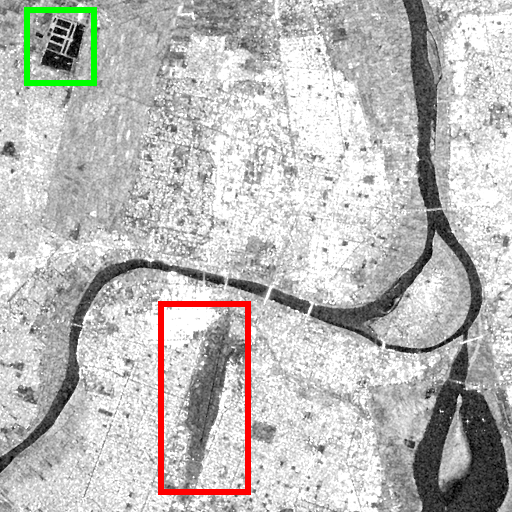}
    \label{fig: intro2}}
    \hfil
    \subfigure[]{\includegraphics[width=0.150\textwidth]{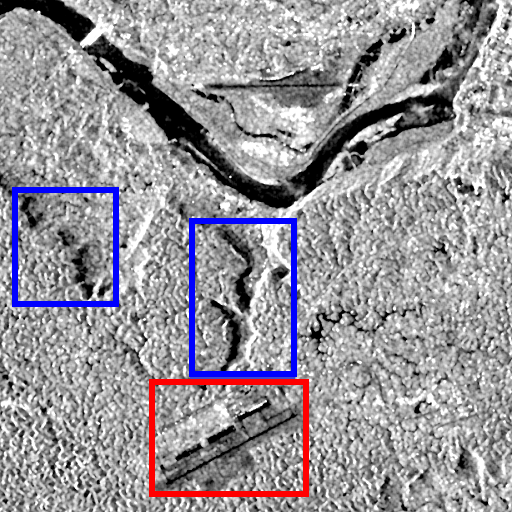}
    \label{fig: intro3}}
    \hfil
    \subfigure[]{\includegraphics[width=0.150\textwidth]{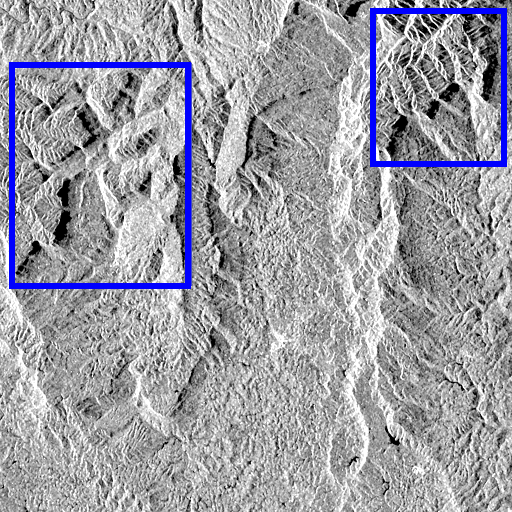}
    \label{fig: intro4}}
    \hfil
    \subfigure[]{\includegraphics[width=0.150\textwidth]{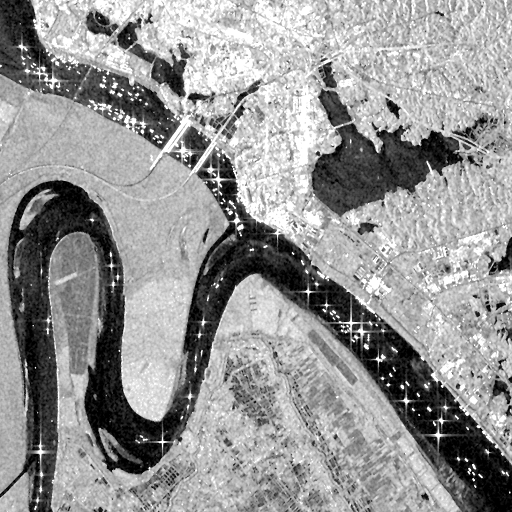}
    \label{fig: intro5}}
    \hfil
    \subfigure[]{\includegraphics[width=0.150\textwidth]{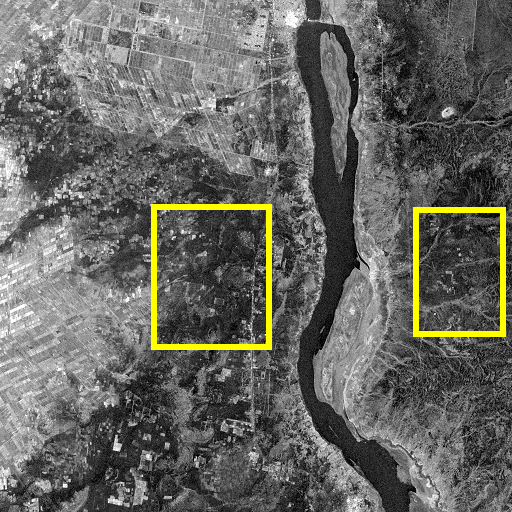}
    \label{fig: intro6}}
    
    \caption{Illustrates several 512 $\times$ 512 patches of Sentinel-1 SAR imagery in our S1GFloods dataset. Detecting changes in SAR data can be challenging due to the complex interactions between the radar signal and the environment.} 
    \label{fig: introchan}
\end{figure*}

Floods are one of the most catastrophic natural disasters, resulting in loss of life, infrastructure destruction, and massive damage to the economy \citep{kundzewicz2013large, zhang2014flood}. In 2021 alone, 206 major flood disasters led to over 29.2 million affected people, 4000 fatalities, and USD 74.6 billion in direct economic globally \citep{report}. This motivates the increasing demand for timely and accurate detection of flooded areas to create rescue plans rapidly.

Synthetic Aperture Radar (SAR) imagery is a popular data source for flooded area detection due to the all-time and all-weather monitoring capability of SAR sensors \citep{huang2020rapid, adriano2019multi}. Based on multi-temporal SAR images, flooded areas are extracted by detecting the changes of water bodies during pre-flood and flooding periods. Inspired by the advance in convolutional neural networks (CNN), various CNN-based change detection methods have been proposed, including FC-Siam-Conc, FC-Siam-Diff \citep{daudt2018fully}, and DASNet \citep{chen2020dasnet}, DTCDSCN \citep{liu2020building}, Siam-NestedUNet \citep{li2020siamese}, and SNUNet-CD \citep{fang2021snunet}. Despite the massive potential for change detection, these methods usually show limited performance in the flood detection task. The main reason for the poor performance is that flooded area detection often needs long-range contextual information. Yet, most CNN-based change detection methods fail to model long-range dependencies due to the inherent locality of convolution operations. As shown in Figure \ref{fig: introchan}, it is difficult to distinguish the flooded area from a local perspective. The local difference in multi-temporal images can be flooded areas or pseudo-change caused by different imaging conditions and registration errors.

To handle this long-range relation learning requirement, the most recent works, such as  BIT \citep{chen2021remote}, SwinSUNet \citep{zhang2022swinsunet}, TransUNetCD \citep{li2022transunetcd} and PA-Former \citep{liu2022pa}, employ vision Transformer (ViT) architecture to capture global interactions and rich contextual information for change detection. Existing ViT-based change detection methods focus on optical images. Compared with optical images, SAR images lack obvious visual perception information such as color and texture. Moreover, due to various land-use categories and the complex terrain, there are many low-intensity areas similar to water bodies in SAR images, e.g., vegetated land, airports, and the hill's shadow (see Figure \ref{fig: introchan}). These easily lead to water-body misrecognition and distinguishing the uninterested changes, such as the change between bare land and the shaded area. In addition to the similar backscatter signal, ubiquitous speckle noise degrading the image quality is another obstacle to accurate flood detection from SAR. 

In response to the performance bottleneck, we aim to explore the ViT-based change detection model for flood detection from SAR images. Inspired by ViTAEv2's \citep{zhang2022vitaev2} success in modeling long-range dependency, we present a differential attention metric-based network (DAM-Net) to achieve global flood detection. As illustrated in Figure~\ref{fig-overall network}, DAM-Net contains two key components, a weight-sharing Siamese backbone based on ViTAEv2 to obtain multi-scale change features of multi-temporal images and tokens containing high-level semantic information of water-body changes, and a temporal-differential fusion module that integrates semantic tokens and change features to generate flood maps. Specifically, the backbone is split into multiple stages. In each stage, we design three modules, i.e., temporal-wise feature extraction (TWFE), cross-temporal change attention (CTCA), and temporal-aware change enhancement (TACE), to extract the change features effectively. In TACE of the last stage, we introduce a class token to record high-level semantic information of water-body changes via the attention mechanism. 

To better train and evaluate the proposed DAM-Net on the global flood detection task, we create a new open-source global-scale flood detection dataset using high-resolution Sentinel-1 SAR images, named \textit{S1GFloods}. Compared with existing related datasets (e.g., SEN12-FLOOD dataset \citep{rambour2020flood}), \textit{S1GFloods} contains more diverse flood events involving various flood causes, such as heavy rains, overflowing rivers, broken dams, tropical storms, and hurricanes. In addition, the proposed datasets have more rich flooded scenes, including wetlands, riverine areas, mountainous regions, urban and rural areas, and vegetation. The diversity and richness of \textit{S1GFloods} are beneficial to develop flood detection methods with excellent performance and good generalization ability. Meanwhile, we annotate the semantic categories of land cover and flooded pixels separately for each image, which allows the researchers to analyze flood events and assess disaster risk further.

In the experiments, we compare DAM-Net with the recent CNN-based and ViT-based methods on the challenging \textit{S1GFloods} dataset. Extensive results demonstrate that DAM-Net outperforms all comparative methods by a large margin in multiple performance metrics, achieving 97.8\% \textit{overall accuracy}, 96.5\% \textit{F1}-score, and 93.2\% \textit{IoU}, respectively.

The main contributions of this paper are summarized as follows:

\begin{itemize}
    \item [1.] We propose a differential attention metric-based network, i.e., DAM-Net, for global flood detection from SAR images, which adopts and modifies the attention module in the vision Transformer to effectively extract semantic tokens and image features of changes within multi-temporal images.
    
    \item [2.] We design a temporal-differential fusion module to explore the relationship between semantic tokens and change features, thereby enhancing differential image features and yielding accurate flood maps.

    \item [3.] We create a global flood detection dataset using multi-temporal Sentinel-1 SAR images, named \textit{S1GFloods}. This dataset contains various flood events and flooded scenes, which allow for developing flood detection methods with good generalization ability.
\end{itemize}

The rest of the paper is organized as follows. Section \ref{Related Work} reviews the related works. Section \ref{Methodology} describes the proposed network in detail. Section \ref{Dataset} introduces the created \textit{S1GFloods} dataset. Section \ref{experiment and analysis} reports experimental results and some discussions. Finally, the conclusion of the paper is drawn in Section \ref{Conclusions}.

\section{Related work}
\label{Related Work}
\subsection{CNN-based Change Detection}
CNN-based change detection methods are widely used for remote sensing images, which can be further divided into two pipelines: two-stage solution and single-stage solution. The former \citep{wang2022fwenet, aparna2022sar} first trains CNN to image segmentation of multi-temporal images, and then compares segmentation masks to generate change regions. This type of method is limited in some applications as they require both change and semantic labels to be available. Moreover, The lack of exploring the relevance of multi-temporal images and accumulation of segmentation errors lead to limited performance. The latter mostly employs Siamese architecture to project multi-temporal images into the same feature space, and then performs feature interaction to produce change results directly. They have shown excellent performance on change detection tasks of common objects (e.g., buildings). For instance, \citet{eftekhari2023building} proposed a dual-branch deep network with a parallel spatial-channel attention mechanism to extract spatial and spectral dependencies, and more discriminative features for detecting building changes. The spatial and channel attention modules enhance the distinction between changed objects and backgrounds. \citet{zhang2022building} applied a multi-scale attention-based change detection model, along with an efficient double-threshold automatic data equalization rule, to address unpredictable change details and the lack of global semantic information. Similarly, \citet{mei2023d2anet} proposed a D2ANet for building localization and multi-level change detection, which includes a dual-temporal aggregation module that excites change-sensitive channels and accurately learns the global change pattern. To capture richer contextual information, \citet{lu2022multi} integrated the features extracted from each layer through a hierarchical structure of bi-temporal images, along with their difference maps, to represent regions of change. The resulting change maps from each layer were then aggregated to improve the effectiveness of change detection. In a similar vein, \citet{pang2023detecting} introduced a multi-tasking guided model that addresses building changes by designing three additional tasks: pixel-wise classification, roof-to-footprint offsets, and identical roof matching flow. 

To boost performance, larger CNN models, optimized loss functions, and attention mechanisms \citep{wang2021ads, yadav2022attentive, zhao2023siam} are widely used to overcome the complexity of remote sensing scenes and the diversity of imaging conditions. However, this group of methods still does not achieve accurate flood detection. Different from building changes that occur in local areas, floods usually span a large spatial range, which leads to flood detection requiring global contextual information. Considering this point, some works adopt technologies, such as dilated convolution, to expand the receptive field. But, the performance improvement is limited due to the inherent locality of convolution operations.
 
\subsection{ViT-based Change Detection}
Since Vision Transformer is effective in modeling long-range dependency, many ViT-based change detection methods have been developed for optical remote sensing images, and report impressive results. For example, \citet{bandara2022transformer} applies several Transformer blocks, consisting of multi-head self-attention and positional encoding modules, to enhance feature representation of bi-temporal optical images. Then, multi-level features are fed into multi-layer perception decoder to yield change maps. To obtain multi-scale long-range context, \citet{liu2022cnn} applies Transformer architecture in conjunction with CNN. Similarly, \citet{li2022transunetcd} uses high-resolution features derived by CNN to address the drawback of ViT of detail loss, thus achieving precise localization of changes. Despite progress, their results are not satisfactory for accurate flood detection based on SAR imagery. The challenges in SAR images, such as scarce visual information, similar backscatter signal, and ubiquitous speckle noise, result in significant performance drops of most existing methods. 

Inspired by the status, recent work \citep{du2022transunet++} has tried to explore the potential of Transformers in change detection with SAR imagery. UNet++ is employed to extract feature representations of multi-temporal SAR images. Then, Transformer encoders followed by an upsampling decoder are used to perform feature interaction and obtain change detection results. Their applicability in change detection of building groups and urban built-up areas has been verified. However, TransUNet++SAR may not be suitable for the flood detection task. The obtained changes of building groups are visually consistent regions with relatively regular boundaries, whereas floods are widely distributed and with complex structures. Consequently, failed flood detection may appear where flood scenes are of complex terrain. Moreover, the CNN-based representation capability of TransUNet++SAR fails to model global context information of each image, which is crucial in flood detection tasks.

\subsection{Flood Detection Datasets}
\label{FDD}
The dataset is an essential prerequisite for developing automatic detection methods, especially deep-learning ones, particularly in flood detection tasks. Several datasets based on SAR imagery have been publicly released. Table~\ref{tab-FD-Datasets} provides a summary of public flood detection datasets. One of the most popular sources is SEN12-FLOOD~\citep{rambour2020sen12}, which combines Sentinel-2 and Sentinel-1 SAR images from four areas that experienced a major flood event during a limited time coverage in 2019. Another dataset, sen1floods11~\citep{bonafilia2020sen1floods11}, consists of Sentinel-1 SAR images acquired after Hurricane Harvey. Similarly, the ETCI-2021~\citep{interagency2021etci} dataset covers five different geographies and includes over $33k$ tiles of Sentinel-1 SAR images. Additionally, the Copernicus Emergency Management Service (CEMS) has produced a dataset consisting of three pairs of Sentinel-1 SAR, COSMO-SkyMed, and RADARSAT2 images, which are located in three regions: Bosnia, Australia, and Scotland during 2022~\citep{yadav2022unsupervised}. Another dataset, RAPID-NRT~\citep{yang2021high}, utilizes the Sentinel-1 SAR archive to provide near-real-time information on 559 flood inundations that occurred over the contiguous United States (CONUS) from 2016 to 2019, using an automated Radar Produced Inundation Diary (RAPID). Recently, MM-Flood~\citep{montello2022mmflood} was introduced, which is the most extensive multimedia dataset tailored for flood identification. It comprises 1,748 Sentinel-1 SAR acquisitions obtained between 2014 and 2021 from 42 countries.

Undoubtedly, these datasets are incredibly valuable. However, we contend that most of them have several shortcomings. Firstly, the vast majority of datasets focus solely on flash floods caused by heavy rainfall during the rainy season and fail to capture other flood events that may occur during the dry season, such as broken dams. Secondly, the data typically comprises a limited number of images and is not adequately geographically distributed to encompass the diversity of the landscape where the flooding occurred, including vegetation, rural, urban, and wetland areas. Thirdly, the current datasets still exhibit an imbalance of sample classes between change and non-change pixels, which can introduce bias in deep learning models towards the majority class, leading to a high rate of false positives. To mitigate these issues, we introduce the \textit{S1GFloods} dataset, which comprises 5,360 pairs of images depicting a wide range of flood types and locations, including urban, rural, and wetland areas across multiple countries and continents, as described in Section~\ref{Dataset}. In comparison to other flood detection datasets, \textit{S1GFloods} is distinguished by its exceptional diversity, generalizability, and balanced class distribution between change and non-change pixels. Its extensive coverage and high-quality data make it a valuable resource for training deep learning models, resulting in improved accuracy and dependability of flood detection systems.

\begin{table*}[hp]
 \centering
 \caption{Compares S1GFloods with other flood detection datasets. Se1, Se2, CSK, and RS2 refer to Sentinel-1, Sentinel-2, COSMO-SkyMed, and RADARSAT2, respectively. The flood types considered in the datasets include Heavy Rain (HR), River Overflow (RO), Broken Dams (BD), Cyclones (C), Tropical Storms (TS), and Hurricanes (H). The symbol $^{*}$ indicates that the event included time series of images.} 
 \renewcommand\arraystretch{1.5}
   \begin{tabular*}{\tblwidth}{@{} LCCCCCCC@{}}
   \hline
   Dataset          & \# Image Pairs & Image Size             & Data Source     & Flood Types     & \# Flood Events &  Acquis. Period & Release \\
   \hline
   MM-Flood         & 1,748          & 2,000 $\times$ 2,000   & Se1             & HR-H            & 42              & 2014-2021 & 2022 \\
   
   SEN12-FLOOD      & 336$^*$       & 512 $\times$ 512        & Se1, Se2        & -               & -               & 2018-2019 & 2020 \\
   
   Sen1Floods11     & 4831          & 512 $\times$ 512        & Se1, Se2        & -               & 11              & 2016-2019 & 2020 \\

   ETCI-2021        & 33,405        & 256 $\times$ 256        & Se1             & HR              & 5               & 2017-2019 & 2021 \\
   
   RAPID-NRT        & 559           & -                       & Se1             & H-C             & 4               & 2016-2019 & 2020 \\
   
   CEMS             & 3             & 25,853 $\times$ 16,748  & Se1-RS2-CSK     & HR              & 3               & 2022 & 2022 \\
   \hline
   S1GFloods (Ours) & 5,360         & 256 $\times$ 256        & Se1             & HR-RO-BD-TS-H-C & 46              & 2015-2022 &  \\
   \hline
   \end{tabular*}%
 \label{tab-FD-Datasets}%
\end{table*}%

\section{Methodology}
\label{Methodology}

\subsection{Network Architecture Overview}
\label{Network Architecture Overview}
Given multi-temporal SAR images captured during pre- and post-flood $\mathbf{I}_{\text{pre}}, \mathbf{I}_{\text{post}} \in \mathbb{R}^{H \times W \times C}$, where $H$, $W$, and $C$ refer to height, width, and channel of the input image, respectively. The flood detection task aims to distinguish the water-body change and generate the binary flood map $\mathbf{M} \in \{0, 1\}^ {H \times W}$. Denoting $1$ as the \textit{change} class (i.e., flooded area) and $0$ as the \textit{non-change} class (i.e., non-flooded area).

\begin{figure*}[ht]
\centering
\includegraphics[width=\textwidth]{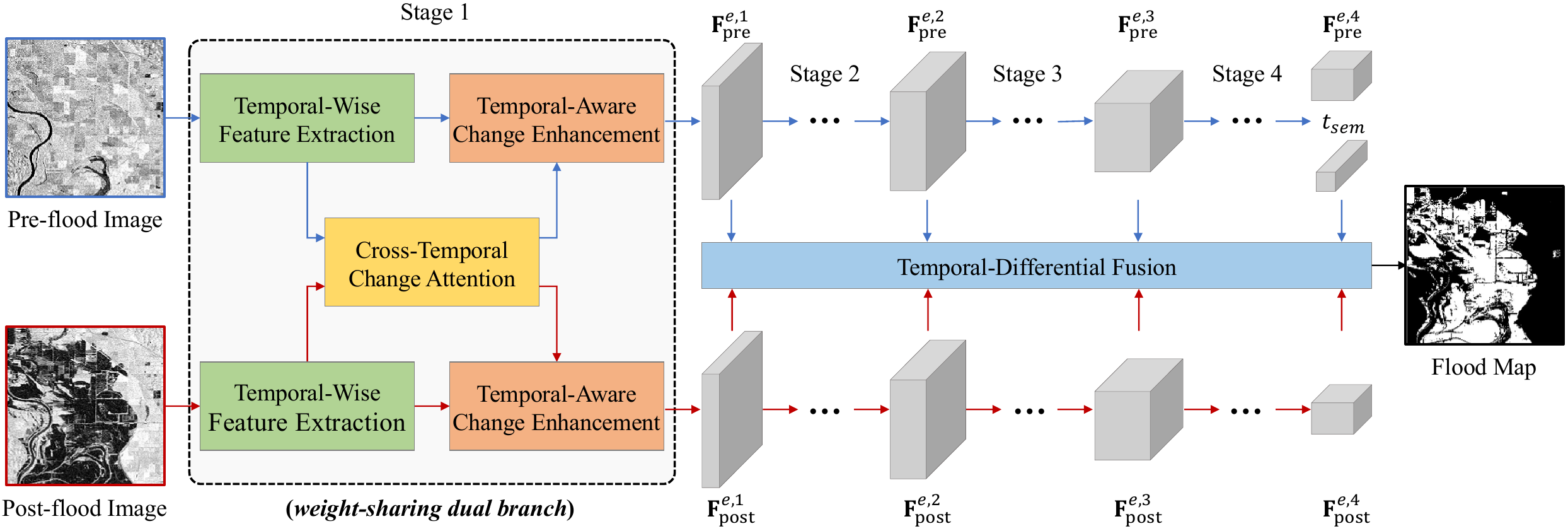}
\caption{The overall architecture of the proposed DAM-Net. Three key modules, i.e., temporal-wise feature extraction, cross-temporal change attention, and temporal-aware change enhancement, are used to perform the representation and interaction of multi-temporal images. Then, a temporal-differential fusion module is adopted for feature integration, yielding the flood map.}
\label{fig-overall network}
\end{figure*}

Figure \ref{fig-overall network} illustrates the overall architecture of the proposed differential attention metric-based network (DAM-Net). DAM-Net takes a weight-sharing Siamese backbone to extract multi-scale change features of multi-temporal SAR images, i.e., $\mathbf{F} _{\text{pre}}^{e,i}$ and $\mathbf{F}_{\text{post}}^{e,i}$ for $i=[1,2,3,4]$. The architecture of ViTAEv2 is adopted to construct the backbone since it is effective for modeling long-range dependency and dealing with scale variance. The backbone contains four stages. In each stage, three key modules, i.e., temporal-wise feature extraction, cross-temporal change attention, and temporal-aware change enhancement, are used to perform the representation and interaction of multi-temporal images, thereby yielding change features. Typically, the flood map then can be derived by combining the change features of multi-temporal images, such as $\boldsymbol{g}(|\mathbf{F}_{\text{pre}}^{e,i}-\mathbf{F}_{\text{past}}^{e,i}|)$ where $\boldsymbol{g}(\cdot)$ indicates the prediction head. However, such a simple procedure is challenging to accurately distinguish the change of interest from the complex and diverse changes. Hence, we introduce a class token in the last stage of the backbone to record the high-level semantic information of water-body changes. Based on the semantic token and high-resolution change features, the flood map is predicted through a temporal-differential fusion module that explores their relationship.

\subsection{Temporal-Wise Feature Extraction}
\label{TWFE}
\begin{figure}[b]
    \centering
    \includegraphics[width=0.3\textwidth]{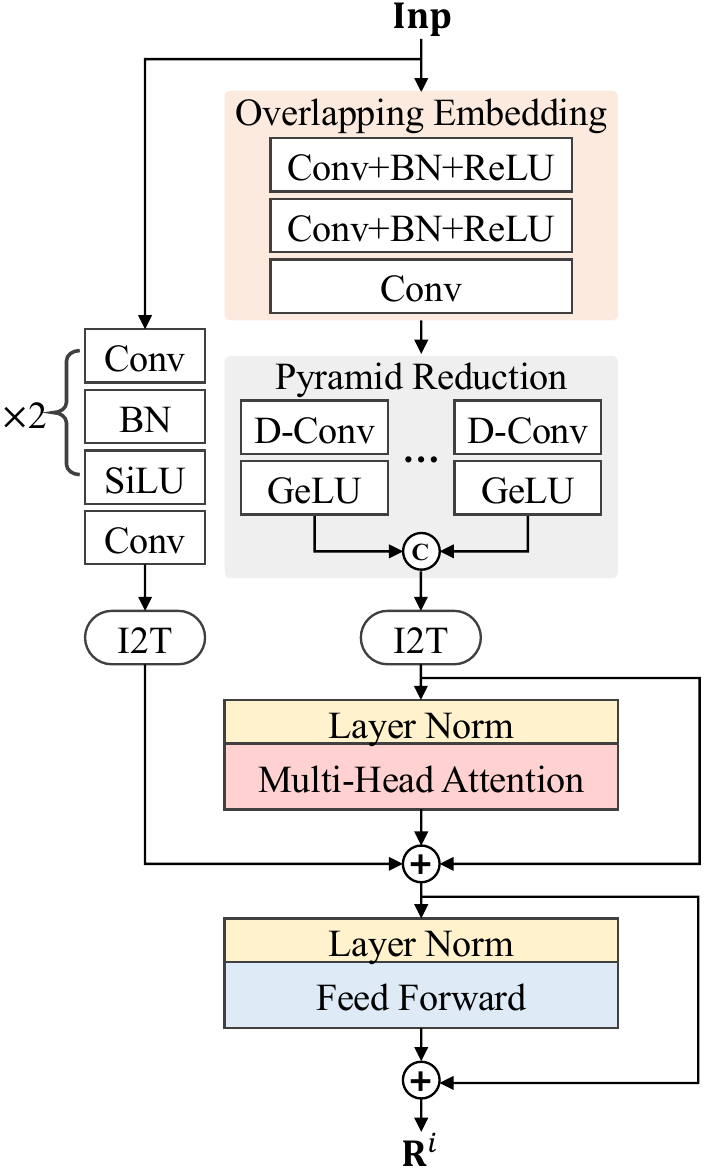}
    \caption{Illustration about the structure of TWFE module.}
    \label{fig-TWFE}
\end{figure}

Long-range contextual information is of great importance for accurate flood detection. Meanwhile, considering the flooded area is of various sizes, flood detection at global scale poses the demand for scale invariance. Since the reduction cell in ViTAEv2 can model long-range dependency and deal with scale variance, we extend it to achieve effective representations of multi-temporal images, which form the temporal-wise feature extraction (TWFE) module. The detailed structure of the TWFE module is demonstrated in Figure \ref{fig-TWFE}. The TWFE module consists of two parallel branches responsible for modeling locality and long-range dependency, followed by a feed-forward network (FFN) for feature transformation. 

In the global dependency branch, instead of directly splitting the input image (or the change features of the previous stage) into non-overlapping patches and then flattening them into token embeddings, we devise an overlapping embedding layer to model local spatial relationships and maintain local continuity between visual tokens. This built-in continuity property boosts the performance of the proposed DAM-Net, which is evidenced in Section \ref{Ablation Study}. Specifically, the overlapping embedding layer is implemented via several convolution operations, as shown in Figure \ref{fig-TWFE}. The convoluted result can then be split and flattened as described above to generate token embeddings for subsequent processes. Notably, we set the convolution operation of kernel size $k=7$, stride $s=2$, and padding $p=3$ for original multi-temporal images, and $k=3$, $s=2$, and $p=1$ for intermediate change features. After the overlapping embedding layer, we follow the design in the reduction cell, adopting a pyramid reduction module (PRM) and a multi-head attention module(MHA) to embed multi-scale information into tokens.

In the local context branch, the combination of an Image2Tokens (I2T) operation and stacked three convolutional layers with two batch normalization layers and two SiLU activation layers in between (named parallel convolutional module, PCM), is used to embed the local context within the tokens. To increase the richness of temporal-wise representation, we fuse tokens generated by the parallel branches and multi-scale context derived by PRM. Then, an FFN and a residual connection are performed to yield the representations of multi-temporal images. Formally, the representations are obtained as follows:
\begin{equation}
\begin{aligned}
& \mathbf{F}_{\text{ms}}^i=\mathrm{PRM}_i(\mathrm{OEL}_i(\mathbf{Inp})), \\
& \mathbf{T}^i=\mathrm{MHA}_i(\mathrm{I2T}(\mathbf{F}_{\text{ms}}^i))+\mathrm{PCM}_i(\mathbf{Inp})+\mathrm{I2T}(\mathbf{F}_{\text{ms}}^i), \\
& \mathbf{R}^i=\mathrm{FFN}_i(\mathbf{T}^i)+\mathbf{T}^i,
\end{aligned}
\end{equation}
where $i$ indicates $i_{th}$ stage or $i_{th}$ TWFE module. $\mathrm{OEL}_i(\cdot)$, $\mathrm{PRM}_i(\cdot)$, $\mathrm{MHA}_i(\cdot)$, $\mathrm{PCM}_i(\cdot)$ and $\mathrm{FFN}_i(\cdot)$ are overlapping embedding layer, pyramid reduction module, multi-head attention module, parallel convolutional module, and feed-forward network. Note that the detailed architecture of the PRM and MHA can refer to \citep{zhang2022vitaev2}. $\mathrm{I2T}(\cdot)$ is a simple operation to reshape feature maps to visual tokens. $\mathbf{F}_\text{ms}^i$, $\mathbf{T}^i$ and $\mathbf{R}^i$ represent multi-scale context, the fusion of parallel branch outputs and the final representation generated by $i_{th}$ TWFE module. The input $\mathbf{Inp}$ can be replaced by the original multi-temporal SAR images and intermediate change features, yielding $\mathbf{R}_\text{pre}^i$ and $\mathbf{R}_\text{post}^i$ ($i=[1,2,3,4]$). As PRM uses stride convolution to reduce the spatial dimension, the resolution of $\mathbf{F}_\text{ms}^i$ ($i=[1,2,3,4]$) are 1/4, 1/8, 1/16, and 1/32 of the original image size. Thus, the representation tokens $\mathbf{R}^i$ ($i=[1,2,3,4]$) are of size $[HW/S_i^2, D_i]$ with downsample rates $S=[4,8,16,32]$. $D=[64,128,256,512]$ is the token dimension.


\subsection{Cross-Temporal Change Attention}
\label{CTCA}
\begin{figure}[!htb]
    \centering
    \includegraphics[width=0.3\textwidth]{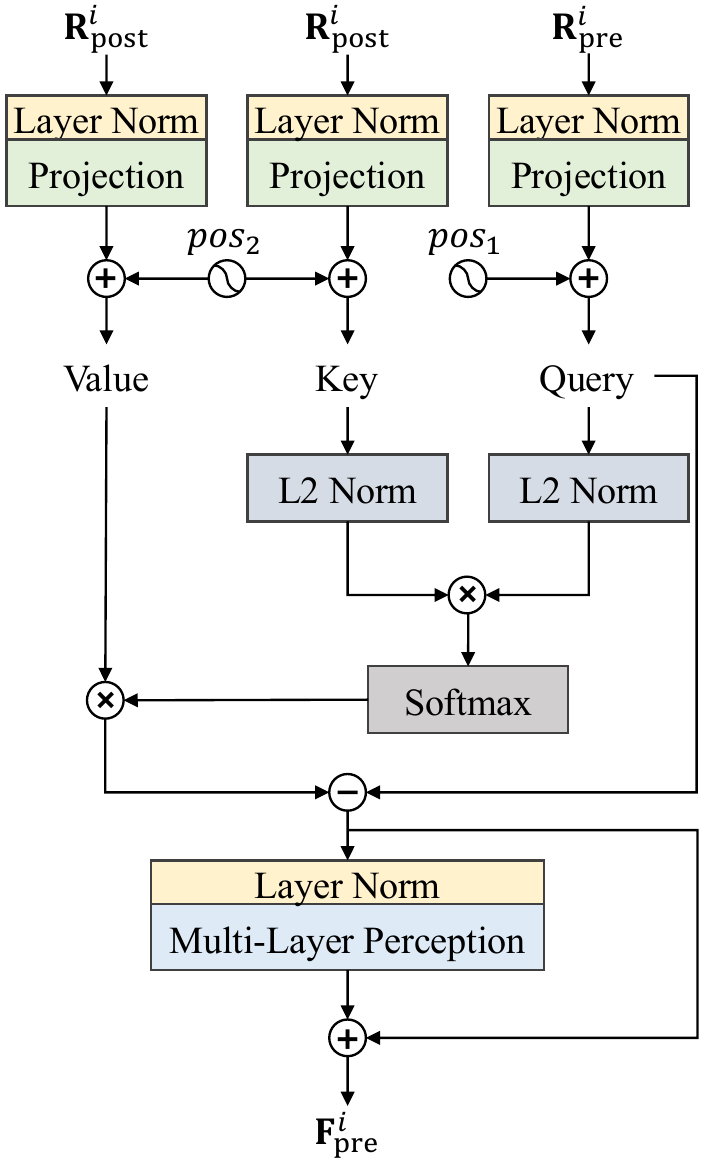}
    \caption{Illustration of the CTCA module.}
    \label{fig-CTCA}
\end{figure}

Based on the representations of multi-temporal SAR images, the cross-temporal change attention (CTCA) module is responsible for exploring the relationship of multi-temporal images and generating effective change information. Figure \ref{fig-CTCA} illustrates the architecture of the proposed CTCA module. Obviously, this module is designed based on the attention mechanism to model long-range dependency between representation tokens of multi-temporal images. This allows the proposed DAM-Net to use long-range contextual information to distinguish changes. For convenience, we take the pre-flood image processing branch as an example to elaborate the CTCA module.

Given the representation tokens $\mathbf{R}_\text{pre}^i$, $\mathbf{R}_\text{post}^i$, we serve $\mathbf{R}_\text{pre}^i$ as Query ($Q$), and $\mathbf{R}_\text{post}^i$ as Key ($K$) and Value ($V$). Specifically, three different projection matrices are first used to perform the transformations. Then, the cross-temporal attention operation is computed as follows:
\begin{equation}
\begin{aligned}
& Q^i,K^i,V^i=\mathbf{R}_\text{pre}^iW_Q^i,\mathbf{R}_\text{post}^iW_K^i,\mathbf{R}_\text{post}^iW_V^i, \\
&\mathbf{CA}^i=Q^i-softmax(\frac{Q^i \cdot {K^i}^T}{\|Q^i\|_2 \cdot\|K^i\|_2}) \cdot V^i,
\end{aligned}
\label{equ-attention}
\end{equation}
where $i$ refers to the $i_{th}$ stage. $W_Q^i$, $W_K^i$, and $W_V^i$ represent the projection matrix for query, key, and value, respectively. $softmax(\cdot)$ is the softmax function. Since the CTCA module aims to capture the change information, we adopt a subtraction operation in Equation \ref{equ-attention}. Moreover, we add a normalization operation via the L2 norm (i.e., $\|\cdot\|_2$) before the dot product of $Q^i$ and $K^i$. The motivation behind such a design is to suppress the influence of noise with the hope that the CTCA module can capture changes more accurately. After the cross-attention, a multi-layer perception (MLP) and a residual connection are employed to generate the change feature $\mathbf{F}_\text{pre}^i$, as follows:  
\begin{equation}
\mathbf{F}_\text{pre}^i=\mathrm{MLP}_i(\mathbf{CA}^i)+\mathbf{CA}^i,
\end{equation}
where $\mathrm{MLP}_i(\cdot)$ is multi-layer perceptron.

For the CTCA module in the post-flood image processing branch, the above process is performed with Query deriving from $\mathbf{R}_\text{post}^i$ and Key, Value obtained through $\mathbf{R}_\text{pre}^i$, finally yielding $\mathbf{F}_\text{post}^i$.

\subsection{Temporal-Aware Change Enhancement}
\label{TACE}
\begin{figure}[hp]
    \centering
    \includegraphics[width=0.3\textwidth]{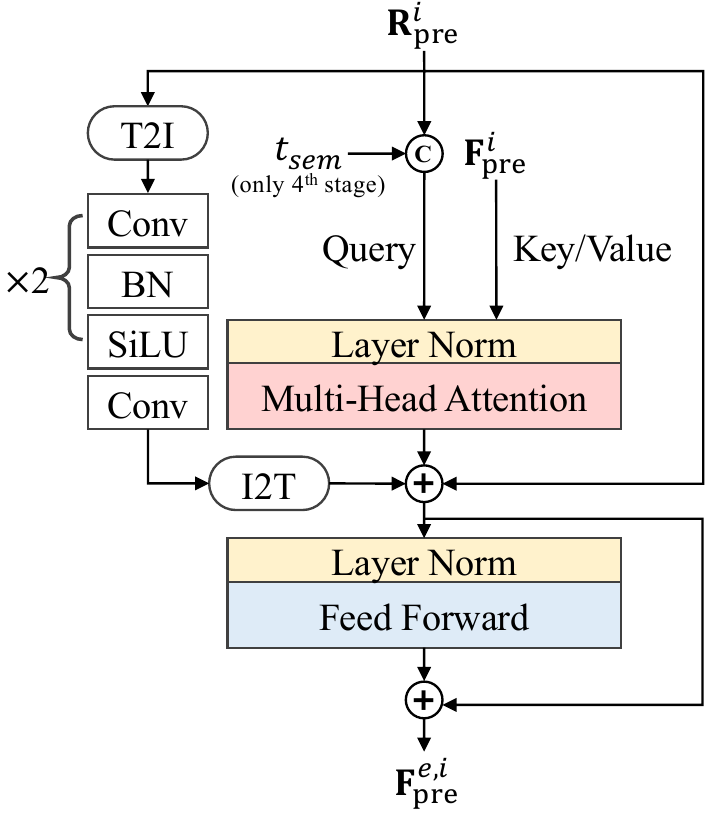}
    \caption{Illustration of the TACE module.}
    \label{fig-TACE}
\end{figure}

Typically, the flood map can be generated by directly integrating change features of multi-temporal images. However, due to different imaging conditions, the simple difference operation between features (see Equation \ref{equ-attention}) may lead to the loss of change information, ultimately hurting performance. To address this issue, we propose to explore the semantic correlation between the representation and change feature of each temporal image, enhancing the change features. Figure \ref{fig-TACE} details the proposed temporal-aware change enhancement (TACE) module.

The TACE module shares a similar architecture with the temporal-wise feature extraction except for the drop of the OEL and PRM. Given the representation of pre-flood SAR images $\mathbf{R}_\text{pre}^i$ (take the pre-flood branch as an example), the PCM and MHA modules are used to process it. Since the TACE module aims to explore the semantic correlation between the representation and change feature, we serve $\mathbf{R}_\text{pre}^i$ as Query and $\mathbf{F}_\text{pre}^i$ as Key, Value in the MHA, different from the TWFE module. Through the attention operation, we hope that features with high correlation in pre-flood representation can be highlighted, thus achieving as complete change information as possible. The generated of the enhanced change feature $\mathbf{F}_\text{pre}^{e,i}$ can be formulated as follows:
\begin{equation}
\begin{aligned}
&\mathbf{T}^i=\mathrm{MHA}_i(\mathbf{R}_\text{pre}^i,\mathbf{F}_\text{pre}^i)+\mathrm{PCM}_i(\mathrm{T2I}(\mathbf{R}_\text{pre}^i))+\mathbf{R}_\text{pre}^i, \\
&\mathbf{F}_\text{pre}^{e,i}=\mathrm{T2I}(\mathrm{FFN}_i(\mathbf{T}^i)+\mathbf{T}^i),
\end{aligned}
\label{equ-tace}
\end{equation}
where $\mathrm{MHA}_i(\cdot,\cdot)$ is a multi-head attention module with the query, key, and value deriving from different entries. $\mathrm{PCM}_i(\cdot)$, $\mathrm{FFN}_i(\cdot)$ represent the PCM and FFN modules, respectively. Note that we ignore the subscript used to distinguish the TWFE and TACE modules when we mention $\mathrm{MHA}_i(\cdot,\cdot)$, $\mathrm{PCM}_i(\cdot)$ and $\mathrm{FFN}_i(\cdot)$, for clarity. $\mathrm{T2I}(\cdot)$ is an operation to reshape tokens back to feature maps. $\mathbf{T}^i$ is the intermediate output. Since the TACE module omit the PRM module, $\mathbf{F}_\text{pre}^{e,i}$ keeps the spatial size of $\mathbf{F}_\text{ms}^i$, i.e., $\mathbf{F}_\text{pre}^{e,i} \in \mathbb{R}^{\frac{H}{S_i} \times \frac{W}{S_i} \times D_i}$. For the post-flood branch, $\mathbf{R}_\text{post}^i$ and $\mathbf{F}_\text{post}^i$ are fed into $i_{th}$ TACE module, yielding the enhanced change feature $\mathbf{F}_\text{post}^{e,i}$. After the sequential processing of four stages, multi-scale change features of multi-temporal SAR images are generated. 

To further accurately identify water-body changes from various changes, we add a class token $t_{sem}$ in $4_{th}$ TACE module of the pre-flood branch, to obtain high-level semantic information of water-body changes, inspired by the success of ViTAEv2 in image classification. Specifically, the class token is concatenated with $\mathbf{R}_\text{pre}^4$ first and then fed into the MHA module as Query. The subsequent process is the same as the above (see Equation \ref{equ-tace}). Consequently, we have the semantic token, which will be used to reduce the interference of uninterested changes (introduced in Section \ref{TDF}). 

\subsection{Temporal-Differential Fusion}
\label{TDF}
\begin{figure}[ht]
    \centering
    \includegraphics[width=0.4\textwidth]{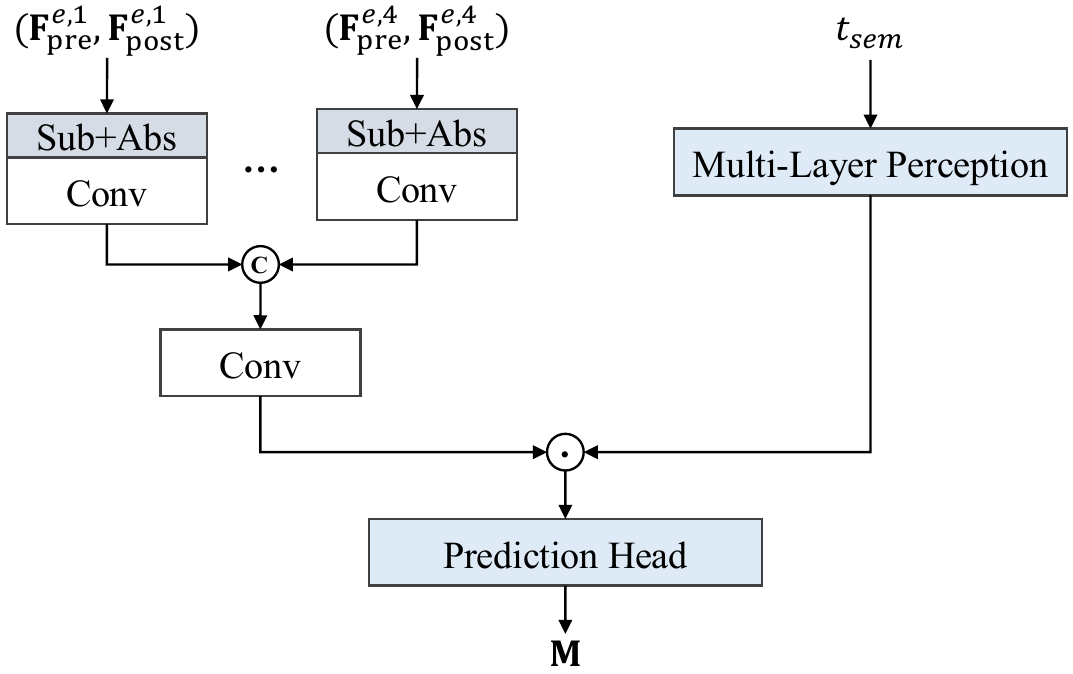}
    \caption{Illustration of the TDF module.}
    \label{fig-TDF}
\end{figure}

The temporal-differential fusion (TDF) module is responsible for exploring the relationship between multi-scale change features and the semantic token, and generating the flood map, as shown in Figure \ref{fig-TDF}. The change features of multi-temporal images share similar change regions, yet the semantic meaning of changes is almost the opposite. For example, from the perspective of the pre-event image, the semantic meaning is that the flood came. Conversely, the semantic meaning is that the flood receded for the post-event image. Motivated by this fact, absolute difference operation is implemented at multiple scales to integrate the change features of multi-temporal images. We name the processing results as temporal-differential features in this paper. The temporal-differential features are fully aware of the change regions and semantic meanings of each multi-temporal image pair, different from the change feature \textit{w.r.t.} single temporal image. 

Then, the multi-scale temporal-differential features are separately fed into the convolutional layers, and fused via a concatenation operation. As the semantic token contains high-level semantic information of water-body changes, we enhance the fused temporal-differential feature $\mathbf{F}_\text{fused}$ with the semantic token, to further suppress the uninterested changes. Finally, we feed the enhanced temporal-differential feature $\mathbf{F}_\text{enhanced}$ to the prediction head to get the flood map $\mathbf{M}$. Formally, the flood map is obtained as follows:
\begin{equation}
\begin{aligned}
&\mathbf{F}_\text{fused}=\mathrm{Concat}[\mathrm{Conv}_i(|\mathbf{F}_\text{pre}^{e,i}-\mathbf{F}_\text{post}^{e,i}|)] \text{ for } i=[1,2,3,4], \\
&\mathbf{F}_\text{enhanced}=\mathrm{Conv}_{1 \times 1}(\mathbf{F}_\text{fused}) \cdot \mathrm{MLP}(t_{sem}), \\
&\mathbf{M}=\sigma(\boldsymbol{g}(\mathbf{F}_\text{enhanced})),
\end{aligned}
\end{equation}
where $\mathrm{Concat}(\cdot)$, $\mathrm{Conv}_i(\cdot)$, $\mathrm{Conv}_{1 \times 1}(\cdot)$, $\boldsymbol{g}(\cdot)$, $\sigma(\cdot)$ are the concatenation operation, the convolutional layer with batch normalization and ReLU function, $1 \times 1$ convolution for adjusting channel dimension, prediction head consisting of the deconvolutional layer and ReLU function, Sigmoid function.

During training, we jointly use the contrastive loss $\mathcal{L}_\text{con}$ and dice coefficient loss $\mathcal{L}_\text {dice}$ to optimize network parameters, considering the class imbalance issue. The total loss $\mathcal{L}_\text{total}$  can be formulated as follows:
\begin{equation}
\begin{aligned}
& \mathcal{L}_\text{total}=\mathcal{L}_\text{con}+\lambda \cdot \mathcal{L}_\text {dice}, \\
& \mathcal{L}_\text{con}=\sum \frac{1}{2}[(1-y)p^2 + y \max(p-m,0)^2], \\
& \mathcal{L}_\text{dice}=1-\frac{2|\mathbf{M} \cap \mathbf{Y}|}{|\mathbf{M}| + |\mathbf{Y}|},
\end{aligned}
\end{equation}
where $p$, $y \in {0,1}$ represent the prediction of the network at a pixel and the corresponding ground truth. $\mathbf{Y} \in {0,1}^{H \times W}$ is the ground truth. $|\mathbf{M}|$ and $|\mathbf{Y}|$ are the number of the predicted change pixel and real change pixel. $|\mathbf{M} \cap \mathbf{Y}|$ denotes the number of the change pixel in the intersection of the two. $m$ is the margin that is enforced for the change pixels. $\lambda$ is the weight of dice coefficient loss. In our experiment, we empirically set $m=1$ and $\lambda=0.4$.

\section{The \textit{S1GFloods} Dataset}
\label{Dataset}
Despite several flood detection datasets, few of them contain sufficient flood events and flood scenes for developing flood detection methods with excellent performance and good generalization ability. For example, the popular flood dataset Sen1Floods11 \citep{bonafilia2020sen1floods11} only contains 11 flood events. To advance the research on flood detection, we introduce \textit{S1GFloods} based on Sentinel-1 SAR imagery. This dataset consists of 5,360 at $256 \times 256$ image pairs, covering 46 heavy flood events between 2015 and 2022, and spanning 6 continents of the world. Specifically,

\textbf{Data Collection.} To achieve diversity, 46 severe flood events spanning 6 continents of the world are collected from ASF\footnote{\url{https://search.asf.alaska.edu/}} Data. Each event has at least one pre-flood image and one post-flood image, as shown in Table~\ref{tab-S1GFlood}. The orbit direction and image size vary by event. The selected events involve the most common and representative types of flood causes, i.e., heavy rain, river overflow, broken dams, tropical storms, and hurricanes.
Such geographic dispersion allows the flood detection method to be aware of various geographic contexts and flood scenes, such as rural areas, mountainous regions, urban areas, vegetated land, rivers, ponds, lakes, and reservoirs. We believe that the performance and generalization ability of flood detection methods can benefit from the diversity of \textit{S1GFloods}.

\begin{figure}[ht]
     \centering
     \subfigure[]{\includegraphics[width=0.1\textwidth]{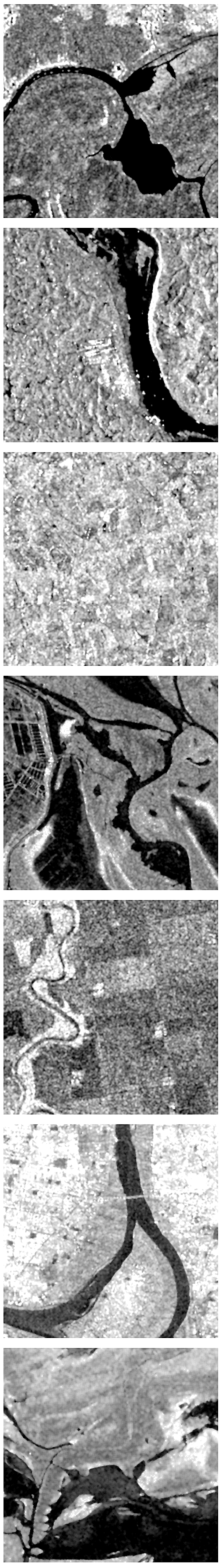}
     \label{fig: samples-patchesa}}
     \hfil
     \subfigure[]{\includegraphics[width=0.1\textwidth]{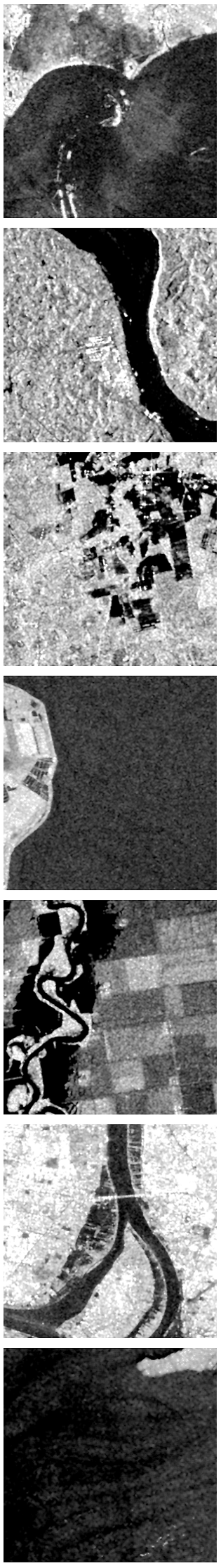}
     \label{fig: samples-patchesb}}
     \hfil
     \subfigure[]{\includegraphics[width=0.1\textwidth]{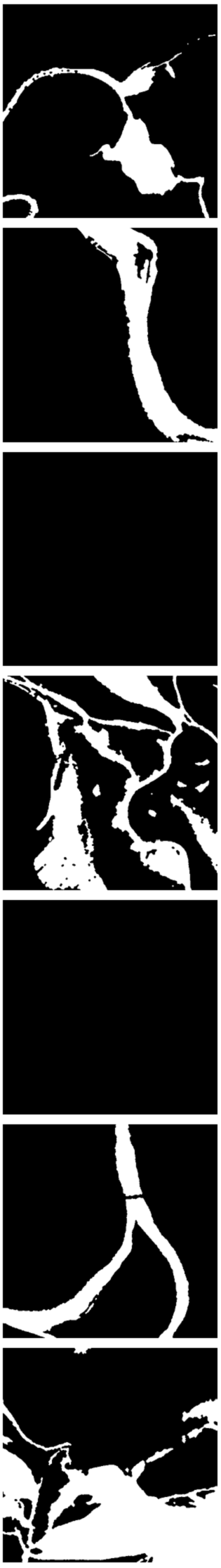}
     \label{fig: samples-patchesc}}
     \hfil
     \subfigure[]{\includegraphics[width=0.1\textwidth]{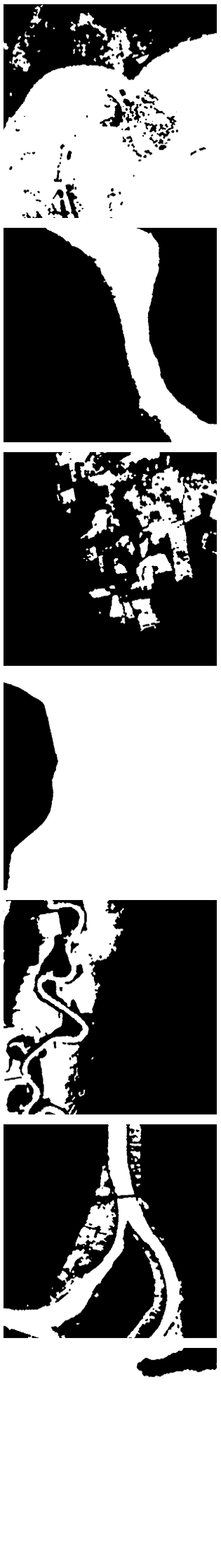}
     \label{fig: samples-patchesd}}
     \caption{Samples from the S1GFloods dataset. The images 1 and 2 are the Sentinel-1 SAR imagery of pre- and post-flood event, displayed in \subref{fig: samples-patchesa} and \subref{fig: samples-patchesb}, respectively. While labels 1 and 2 are the corresponding annotation maps which indicate the flooded areas with pixel-precise, showed in \subref{fig: samples-patchesc} and \subref{fig: samples-patchesd}.}
     \label{fig: samples-patches}
\end{figure}

 \textbf{Data Annotation.} For each event, we first distinguish regions affected by the flood from raw imagery. Then, these regions are cropped into non-overlapping image patches with the size of $256 \times 256$. In total, 5,360 image pairs are obtained. During the data annotation process, we adopted a semi-automated labeling strategy to generate ground truth of flood. Specifically, we first applied back-scatter coefficient histogram analysis provided by python-snappy of the European Space Agency \citep{s1tbx}, to segment water masks. The segmentation threshold $\sigma_0 = -18$ dB for all image pairs. Since other objects such as shadows, airports, and vegetated land, are shown as dark areas like water, automatic annotation inevitably has many errors. Hence, manual annotation was introduced to refine the generated masks. Based on high-resolution optical satellite imagery from Google Earth\footnote{\url{https://earth.google.com/}} and Maxar Technologies\footnote{\url{https://www.maxar.com/}}, skilled remote-sensing image interpreters can distinguish between water and interference objects, then correct false mask. Then, we identified flood and permanent water by comparing water masks of each image pair, and performed boundary smoothing and isolated noise removal via morphological operations (i.e., erosion and dilation). Figure~\ref{fig: samples-patches} shows examples of multi-temporal SAR images and ground truth.

\textbf{Training, Validation, and Testing Data Division.} The \textit{S1GFloods} Dataset aims to achieve real-world flood issue diversity, thereby advancing the development of effective flood detection. Hence, we consider not only the number of images but also the images to be sufficiently representative for different flood events and scenes, during data division. Namely, training, validation, and testing datasets should contain all flood causes and scenes. Following this scheme, we have 4,300, 530, and 530 image pairs for training, validation, and testing, respectively. To better evaluate the practicality of the flood detection method, we add two large-sized image pairs only for flood inundation mapping testing. These images are collected from Nebraska and Iran, with the size of $12705 \times 17952$ or $9840 \times 5276$.

\begin{table*}[htp]
  \centering
  \caption{Detailed descriptions of global flood events in the \textit{S1GFloods} dataset. The symbol $^*$ indicates that the event included multiple images of post-event flooding. Events labeled with $^{**}$ signify the inclusion of an external test set.}
  \renewcommand\arraystretch{1.17}
    \begin{tabular*}{\tblwidth}{@{} LLLCCCL@{}}
    \hline
    Continent                           & Location                  & Event Date        & Image Size    & Absolute Orbit & Path & Flood Cause \\
    \hline
    \multirow{15}[2]{*}{Asia}           & Sylhet, Bangladesh        & 07/12/2017        & 10768 × 12584 & 17438 & 041 & Heavy Rain \\
                                        & Sunamganj, Bangladesh     & 06/23/2022        & 11087 × 10456 & 43797 & 150 & Heavy Rain \\
                                        & Chongqing, China          & 08/14/2020        & 15259 × 07663 & 33902 & 055 & Heavy Rain \\
                                        & Nanchang, China           & 07/14/2020        & 13080 × 14681 & 22466 & 040 & Heavy Rain \\
                                        & Wuhan, China              & 07/13/2020        & 13135 × 11214 & 33435 & 113 & Heavy Rain \\
                                        & Assam, India              & 05/18/2022        & 09874 × 10934 & 43265 & 143 & Heavy Rain \\
                                        & Goalpora, India           & 05/16/2022        & 07765 × 06634 & 43236 & 114 & Heavy Rain \\
                                        & Baluchestan, Iran         & 01/13/2020$^{**}$ & 09840 × 05276 & 19785 & 159 & Cyclone Shaheen \\
                                        & Sistan, Iran              & 01/7/2020         & 03009 × 02405 & 30783 & 086 & Cyclone Shaheen \\
                                        & Kumamoto, Japan           & 06/28/2020$^{*}$  & 07113 × 05436 & 22232 & 156 & Tropical Storm \\
                                        & Attapeu, Laos             & 07/25/2018$^{*}$  & 09987 × 09584 & 22950 & 128 & Broken Dam \\
                                        & Shahdadkot, Pakistan      & 08/30/2022        & 11478 × 10235 & 44775 & 078 & Heavy Rain \\
                                        & Qambar, Pakistan          & 08/30/2022        & 07698 × 04538 & 44775 & 078 & Heavy Rain \\
                                        & Kastamonu, Turkey         & 08/11/2021        & 05482 × 02749 & 28193 & 167 & Heavy Rain \\
                                        & Bartin, Turkey            & 08/12/2021        & 03589 × 04619 & 39184 & 087 & Heavy Rain \\
    \hline
    \multirow{2}[2]{*}{Europe}          & Duisburg, Germany         & 07/18/2021        & 01574 × 02500 & 38835 & 088 & Heavy Rain \\
                                        & Roermond, Netherlands     & 07/18/2021        & 02169 × 02385 & 38835 & 088 & Heavy Rain \\
    \hline
    \multirow{2}[2]{*}{Oceania}         & Port Macquarie, Australia & 03/19/2021$^{*}$  & 06933 × 04770 & 37071 & 074 & Heavy Rain \\
                                        & Sydney, Australia         & 03/24/2021        & 05612 × 03389 & 37071 & 074 & Heavy Rain \\
    \hline
    South America                       & Bahia, Brazil             & 12/14/2021        & 06408 × 03590 & 41002 & 155 & Heavy Rain \\
    \hline
    \multirow{5}[2]{*}{North America}   & Florence, America         & 09/19/2018        & 10404 × 12279 & 23774 & 077 & Hurricane \\
                                        & Grafton, America          & 04/22/2019        & 07923 × 04119 & 26909 & 062 & Hurricane \\
                                        & Omaha, Nebraska           & 03/18/2019$^{**}$ & 12705 × 17952 & 26385 & 063 & Hurricane \\
                                        & Iowa, Nebraska            & 03/16/2019        & 01817 × 01848 & 26356 & 034 & Hurricane \\
                                        & Lakota, North Dakota      & 05/28/2019        & 08966 × 13417 & 16437 & 136 & Hurricane \\
    \hline
    \multirow{23}[2]{*}{Africa}         & Khartoum, Sudan           & 09/23/2020        & 08917 × 10318 & 23497 & 021 & Heavy Rain \\
                                        & Mayom, South Sudan        & 09/11/2019        & 04310 × 03489 & 18845 & 094 & Heavy Rain \\
                                        & Pibor, South Sudan        & 05/11/2019        & 03759 × 04027 & 29770 & 123 & Heavy Rain \\
                                        & Yirol, South Sudan        & 04/11/2019        & 02285 × 04037 & 18772 & 021 & Heavy Rain \\
                                        & Beira, Mozambique         & 01/25/2021$^{*}$  & 08361 × 09306 & 36296 & 174 & Tropical Storm \\
                                        & Mavozaza, Madagascar      & 01/31/2020$^{*}$  & 08922 × 06729 & 31038 & 166 & Heavy Rain \\
                                        & Arusha, Tanzania          & 04/24/2020        & 05509 × 04376 & 21280 & 079 & Heavy Rain \\
                                        & Moshi, Tanzania           & 04/28/2020        & 03589 × 02793 & 32329 & 057 & Heavy Rain \\
                                        & Hayq, Ethiopia            & 08/28/2020        & 08033 × 09371 & 34101 & 079 & Heavy Rain \\
                                        & Timbuktu, Mali            & 12/13/2022        & 05590 × 04331 & 46317 & 045 & Heavy Rain \\
                                        & Dire, Mali                & 08/15/2021$^{*}$  & 03995 × 05006 & 39244 & 147 & Heavy Rain \\
                                        & Chikwawa, Malawi          & 03/07/2019        & 07599 × 08694 & 26226 & 079 & Heavy Rain \\
                                        & Rundu, Namibia            & 02/17/2020$^{*}$  & 09094 × 08926 & 20311 & 160 & Heavy Rain \\
                                        & Agadez, Niger             & 08/14/2021        & 05118 × 04908 & 39229 & 132 & Heavy Rain \\
                                        & Tahoua, Niger             & 08/19/2021        & 04317 × 04743 & 39302 & 030 & Heavy Rain \\
                                        & Nema, Mauritania          & 09/11/2020$^{*}$  & 10337 × 09804 & 34315 & 118 & Heavy Rain \\
                                        & Masvingo, Zimbabwe        & 01/24/2021$^{*}$  & 05592 × 05099 & 36274 & 152 & Heavy Rain \\
                                        & Ndola, Zambia             & 03/17/2020$^{*}$  & 07931 × 09834 & 31695 & 123 & Heavy Rain \\
                                        & Lokoja, Nigeria           & 10/13/2022$^{*}$  & 04459 × 03369 & 45427 & 030 & Heavy Rain \\
                                        & Beledweyne, Somalia       & 04/25/2018$^{*}$  & 04765 × 07894 & 10642 & 116 & Overflowing River \\
                                        & Bardere, Somalia          & 04/29/2018$^{*}$  & 03375 × 02795 & 21684 & 087 & Overflowing River \\
                                        & Wadi El Natrun, Egypt     & 10/28/2015        & 06983 × 04805 & 08355 & 058 & Heavy Rain \\
                                        & Ras Gharib, Egypt         & 10/28/2016$^{*}$  & 09653 × 08539 & 02709 & 058 & Heavy Rain \\
    \hline
    \end{tabular*}%
  \label{tab-S1GFlood}%
\end{table*}%

\section{Experiment and Analysis}
\label{experiment and analysis}

\subsection{Experiment Setting}
\subsubsection{Comparative Methods}
We compare the proposed DAM-Net against the state-of-the-art methods, originally designed for change detection of remote sensing images, on the considered global flood detection task. Specifically, we report the results of five CNN-based and two ViT-based methods. These methods are as follows:

FC-Siam-Diff \citep{daudt2018fully}: adopts a weight-sharing Siamese backbone based on UNet encoder to extract multi-scale features of bi-temporal images. Then, the feature differences of bi-temporal images are concatenated with multi-scale image features to detect changes via UNet decoder.

FC-Siam-Conc \citep{daudt2018fully}: shares a similar architecture with FC-Siam-Diff except for the concatenation operation. This method directly concatenates multi-scale features of bi-temporal images in the decoder.

DTCDSCN \citep{liu2020building}: introduces auxiliary semantic segmentation tasks and design the spatial-channel attention to learn more discriminative features. Meanwhile, they employ the spatial pyramid pooling to enlarge the receptive field of CNN, thus capturing more context information. 

SNUNet-CD \citep{fang2021snunet}: combines Siamese network and NestedUNet to alleviates the information loss in the deep change detection network. In addition, they use channel attention mechanism to address semantic gaps and localization differences in multi-stage output integration.

Siam-NestedUNet \citep{li2020siamese}: explore the performance of different interactions (i.e, concatenation, subtraction and the combination of them) between two Siamese branches based on the combination of Siamese network and NestedUNet. In this paper, we adopt Siam-NestedUNet with the concatenation operation for the comparisons since this architecture achieves the best performance as reported in \citep{li2020siamese}.

BIT \citep{chen2021remote}: first uses CNN backbone to extract features of bi-temporal images and expresses them into semantic tokens. Then, the Transformer encoder and decoder are introduced to model contexts and refine the original features,respectively. Finally, the absolute the subtraction of the two refined features is fed into a shallow FCN for change map generation. In this paper, we use BIT with ResNet-50 \citep{he2016deep} as backbone for comparisons.

SwinSUNet \citep{zhang2022swinsunet}: is a pure transformer network with Siamese U-shaped structure to solve change detection problems. It uses Swin transformer blocks \citep{zhang2022swinsunet} as basic units to construct encoder for multi-scale feature extraction, fusion for feature merger of bi-temporal features, and decoder for recovering change information. 

\subsubsection{Implementation details}
We trained all models using the PyTorch framework on an NVIDIA GRID RTX8000 (8GB) GPU. Note that we use public codes of the comparative methods for comparison, to ensure fairness. Concretely, for five CNN-based methods and BIT, we set the initial learning rate to $e^{-3}$, weight decay to $5e^{-4}$ and use stochastic gradient decent (SGD) \citep{qian1999momentum} with the momentum of $0.99$ to optimize the model. While for SwinSUNet and DAM-Net, the initial learning rate and weight decay are $e^{-2}$, $6e^{-5}$. The AdamW algorithm \citep{kingma2014adam} with $\beta1$ of $0.900$, and $\beta2$ of $0.999$ is used to optimize the model. All models are with 100 epochs and batch size of 8. During training, the learning rate will linearly decay when the epoch reaches 20, 50 and 90. Image transformations, such as rotation and mirror, are adopted to augment the training data. After training, we select the model with best performance on the validation set for comparisons on the testing set.

\subsubsection{Performance Metrics}
To quantitatively assess the performance of DAM-Net, we report the evaluation of precision (P), recall (R), F1-score (F1), overall accuracy (OA) and intersection over union (IoU). These metrics are defined as follows:
\begin{equation}
   \begin{aligned}
       &\text{P}=\frac{\text{TP}}{\text{TP}+\text{FP}}, \\
       &\text{R}=\frac{\text{TP}}{\text{TP}+\text{FN}}, \\
       &\text{F1}=2 \times \frac{\text{P} \times \text{R}}{\text{P}+\text{R}}, \\
       &\text{OA}=\frac{\text{TP}+\text{TN}}{\text{TP}+\text{TN}+\text{FN}+\text{FP}}, \\
       &\text{IoU}=\frac{\text{TP}}{\text{TP}+\text{FN}+\text{FP}},
   \end{aligned} 
\end{equation}
where true positive (TP), true negative (TN) represent the number of the change and unchange pixel predicted correctly. while false positive (FP), false negative (FN) indicate the number of the unchange and change pixel predicted incorrectly.

\subsection{Comparison with the State-of-the-art}
\label{Comparision}
Table \ref{tab-comparision} reports the performance of different change detection methods on the \textit{S1GFloods} dataset. From Table \ref{tab-comparision}, we can observe that FC-Siam-Diff performs worst within all CNN-based methods. DTCDSCN, SNUNet-CD, and Siam-NestedUNet gradually improve performance through larger receptive field, adopting attention mechanism, and more dense connections. SwinSUNet and BIT achieve competitive precision (nearly 94\%) without sophisticated structures such as NestedUNet in Siam-NestedUNet and SNUNet-CD, thus demonstrating the importance of modeling long-range dependency for flood detection. However, it is worth noting that these two methods are not competitive with the CNN-based methods on recall (inferior to 94\% vs most exceeding 95\%). The main reason is that the CNN-based methods are expert in modeling locality which are important for the boundary and small objects, and perform multi-scale feature integration to address information loss. This observation inspires us to explore the potential of ViTAEv2 in change detection since ViTAEv2 can model locality and long-range dependency. Consequently, our method achieves a comparable recall (94.7\%), and the highest score on precision (96.0\%) and F1 (95.3\%), with only 29\% FLOPs of Siam-NestedUNet (sub-optimal performance). This confirms that DAM-Net has an excellent trade-off between detection performance and model complexity.
\begin{table*}[hp]
\centering
\caption{Comparison with state-of-the-art methods on the \textit{S1GFloods} dataset. The best result is highlighted in \textbf{bold}. Params., FLOPs refer to the number of network parameters and floating point operations, used to evaluate the complexity of the model.}
\renewcommand\arraystretch{1.2}
\begin{tabular*}{\tblwidth}{@{} LCCCCCC@{}}
\hline
\multicolumn{1}{c}{Method} & Architecture & \multicolumn{1}{c}{Params. (M)} & \multicolumn{1}{c}{FLOPs (G)} & \multicolumn{1}{c}{P (\%)} & \multicolumn{1}{c}{R (\%)} & \multicolumn{1}{c}{F1 (\%)} \\
\hline
FC-Siam-Diff \citep{daudt2018fully}   & \multirow{5}{*}{CNN} & 1.4 & 9.4 & 86.6 & 94.7 & 90.2 \\
FC-Siam-Conc \citep{daudt2018fully}   &                      & 1.5 & 10.6 & 89.4 & 94.6 & 91.7 \\
DTCDSCN \citep{liu2020building}       &                      & 31.3 & 26.4 & 89.3 & \textbf{95.8} & 92.3 \\
SNUNet-CD \citep{fang2021snunet}      &                      & 12.0 & 109.6 & 93.8 & 95.0 & 94.3 \\
Siam-NestedUNet \citep{li2020siamese} &                      & 12.0 & 109.6 & 94.6 & 95.3 & 94.9 \\
\hline
SwinSUNet \citep{zhang2022swinsunet}  & \multirow{3}{*}{ViT} & 28.2 & 31.3 & 94.0 & 92.8 & 93.4 \\
BIT \citep{chen2021remote}            &                      & 24.4 & 25.2 & 93.9 & 93.5 & 93.7 \\
DAM-Net (Ours)                        &                      & 19.5 & 32.0 & \textbf{96.0} & 94.7 & \textbf{95.3} \\
\hline
\end{tabular*}
\label{tab-comparision}
\end{table*}

\begin{figure*}[hp]
     \centering
     \includegraphics[width=0.798\textwidth]{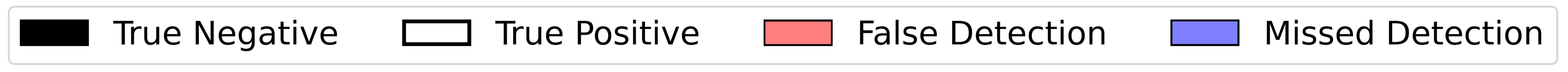}
     \\
     \includegraphics[width=0.998\textwidth]{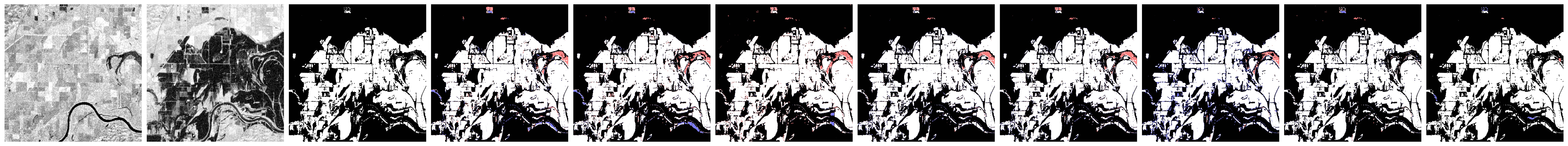}
     \\
     \includegraphics[width=0.998\textwidth]{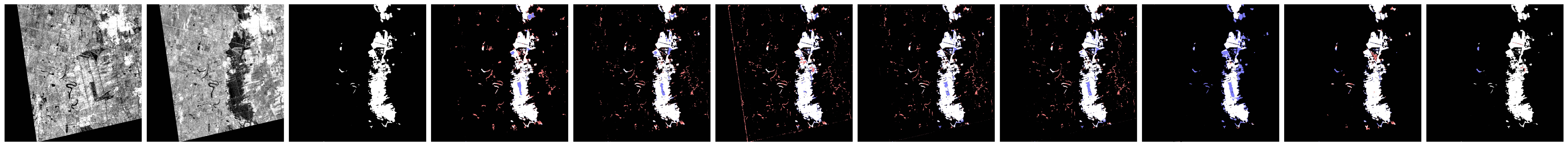}
     \\
     \includegraphics[width=0.998\textwidth]{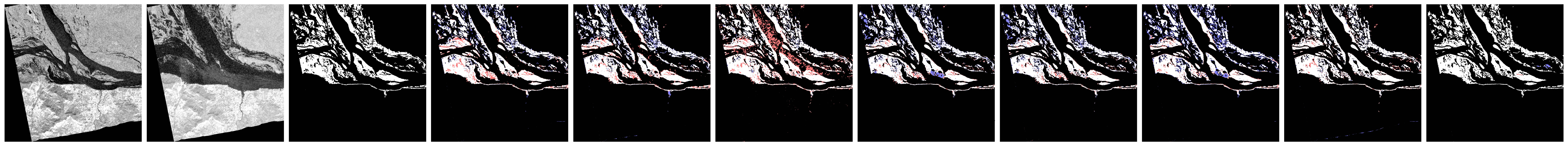}
     \\
     \includegraphics[width=0.998\textwidth]{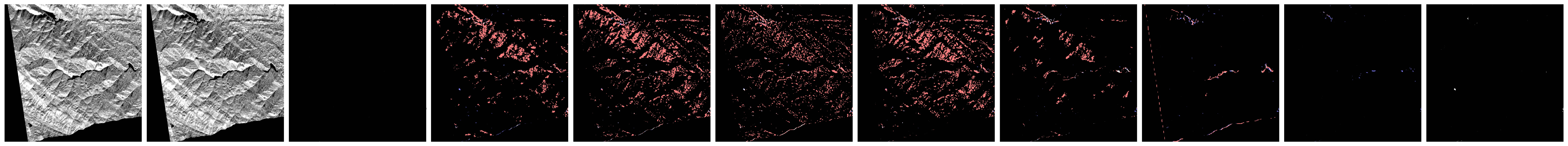}
     \\
     \includegraphics[width=0.998\textwidth]{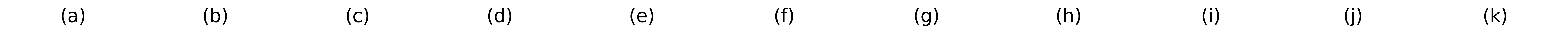}      
     \caption{Visualization results of the comparison. (a), (b), and (c), represent the pre-flood image, post-flood image, and binary ground truth, respectively. (d)-(k) are flood detection results of FC-Siam-Diff, FC-Siam-Conc, DTCDSCN, SNUNet-CD, Siam-NestedUnet, SwinSUNet, BIT, and (k) DAM-Net, respectively.}
     \label{fig:quantitative}
\end{figure*}

In addition to quantitative results, we visualize some representative detection results for intuitive comparisons, as shown in Figure~\ref{fig:quantitative}. Different colors are used to represent true positive (white), true negative (black), false detection (red), and missed detection (blue). The first three multi-temporal SAR image pairs show vegetated land (Iowa-Nebraska), urban areas (Sistan-Iran), and regions along rivers (Lokoja-Nigeria) hit by floods, respectively. For these scenes, flood detection is susceptible to interferences caused by agricultural irrigation, complex land-use categories, and seasonal changes in river runoff. The last image pair was collected from the Chongqing mountainous area of China, which did not suffer a flood event. However, the changes of the hill's shadow caused by the variations in viewpoint, orbit direction, etc., are easily mistaken as flooded areas. Because both the shadow and water bodies display as nearly dark areas. Despite many challenges, the flood map generated by DAM-Net has complete details and is closer to the ground truth than the comparative methods. These results confirm the effectiveness of DAM-Net on the flood detection task.

\begin{figure*}[!ht]
    \centering
    \subfigure[]{\includegraphics[width=0.995\textwidth]{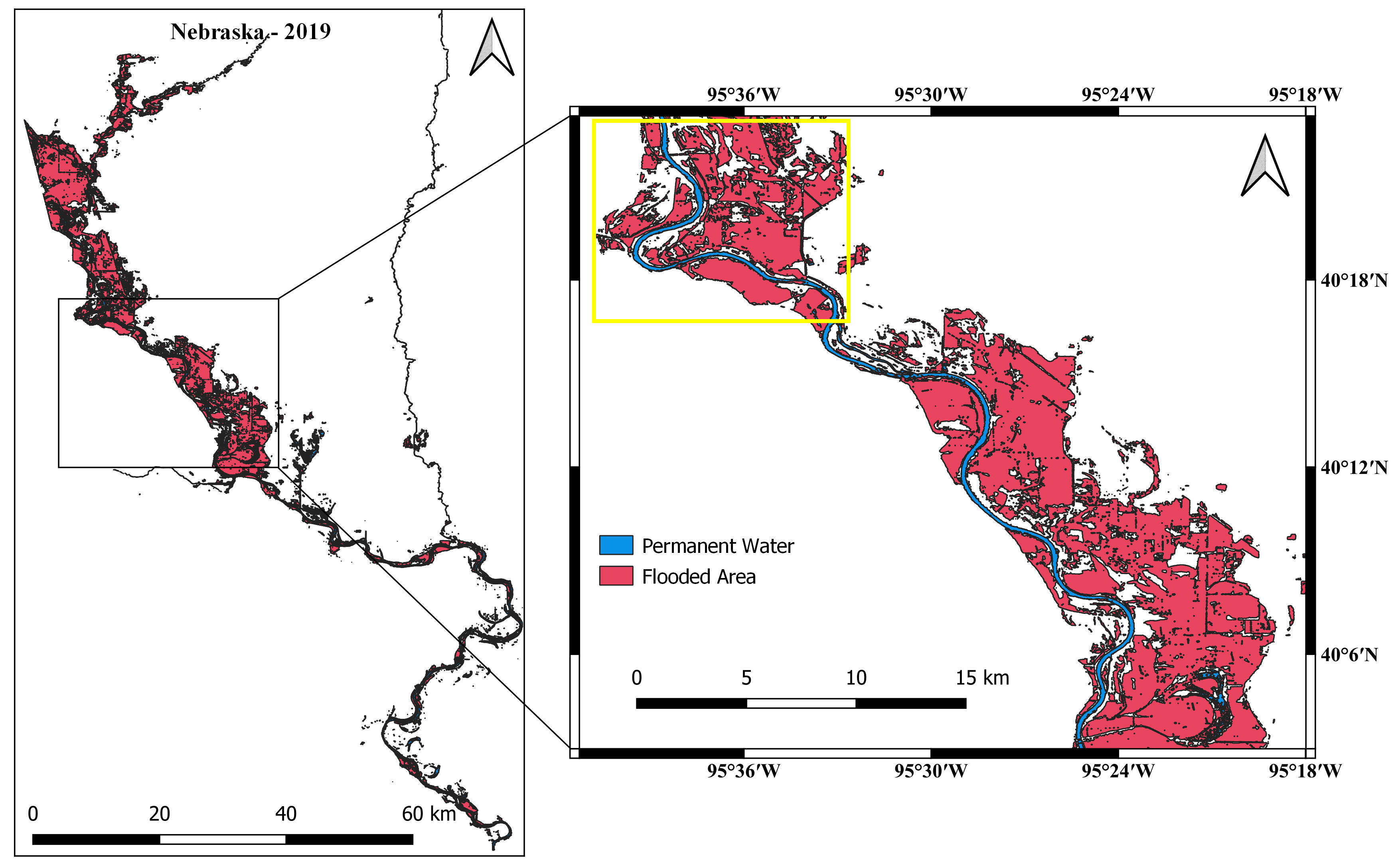}
    \label{nebraskaa}}
    \\
    \subfigure[]{\includegraphics[width=0.32\textwidth]{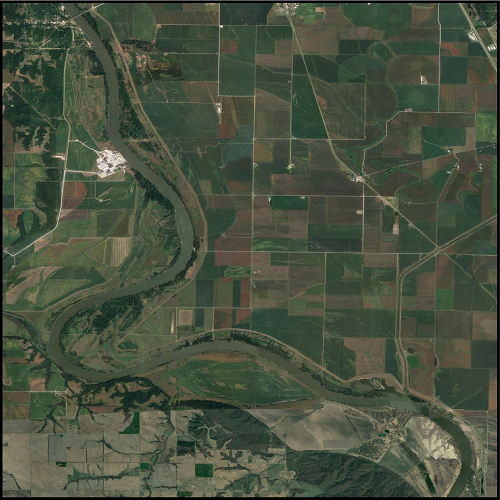}
    \label{nebraskab}}
    \hfil
    \subfigure[]{\includegraphics[width=0.32\textwidth]{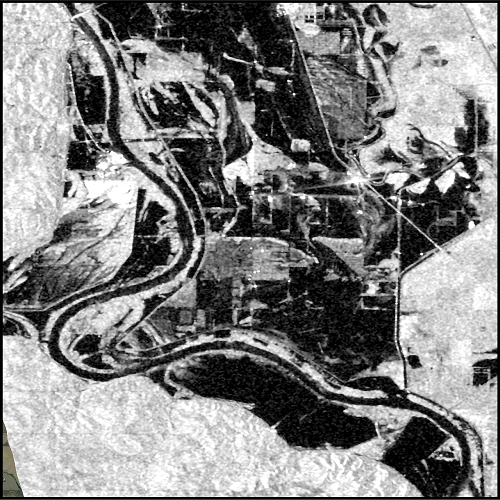}
    \label{nebraskac}}
    \hfil
    \subfigure[]{\includegraphics[width=0.32\textwidth]{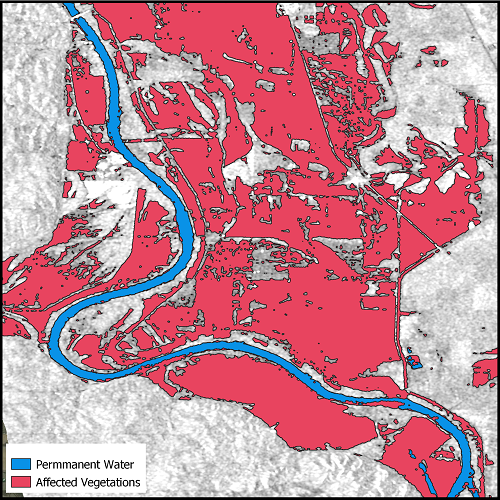}
    \label{nebraskad}}
    
    \caption{Flood inundation mapping of the Nebraska region. \subref{nebraskaa} The Nebraska region has a total area of 613.581 sq km, and the flooding occurred on March 18, 2019. \subref{nebraskab}-\subref{nebraskad} provide zoomed-in views of the pre-flood optical image, post-flood SAR image, and flood mapping of yellow box in \subref{nebraskaa}.}
    \label{testingmap1}
\end{figure*}

\begin{figure*}[!ht]
    \centering
    \subfigure[]{\includegraphics[width=0.995\textwidth]{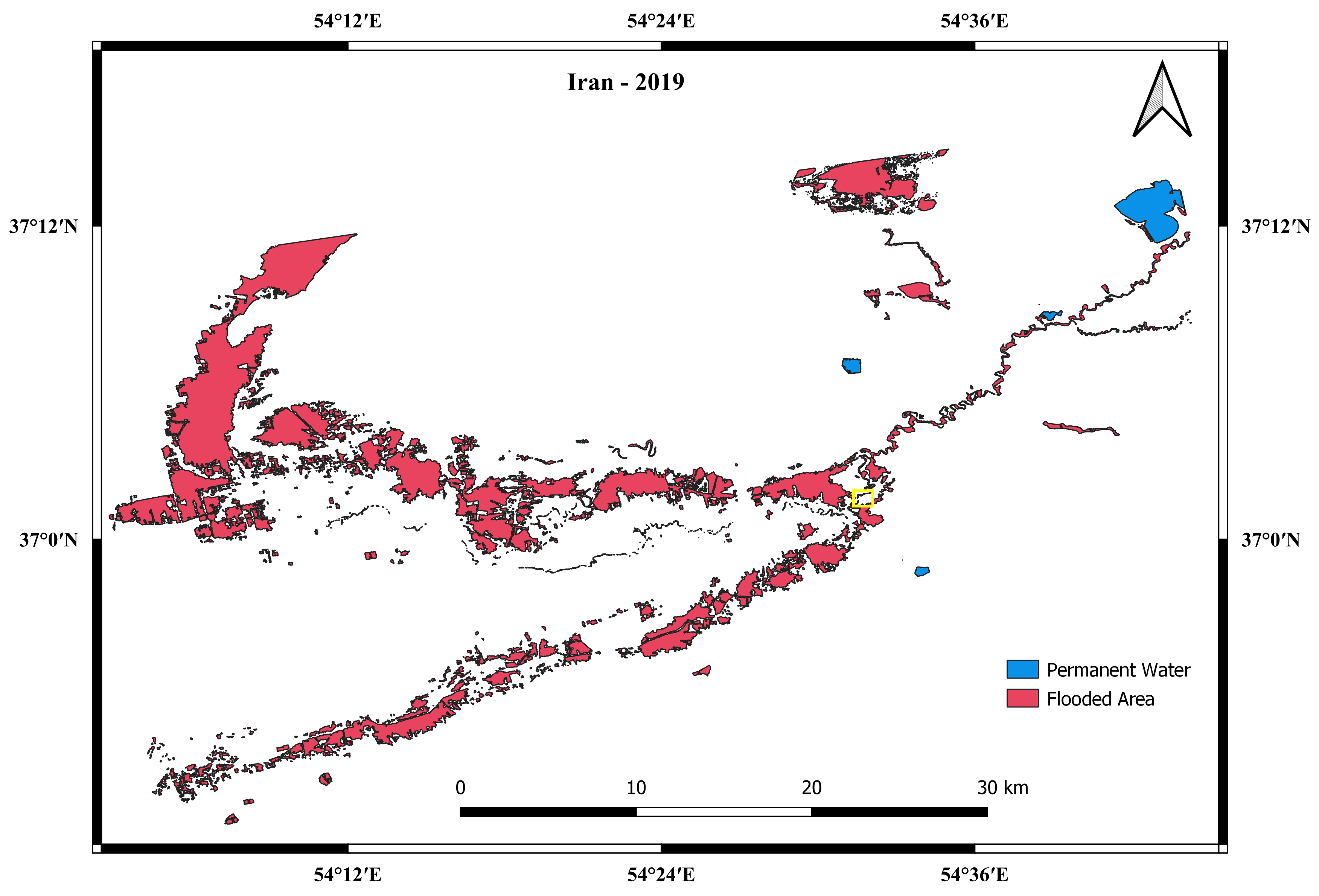}
    \label{irana}}
    \\
    \subfigure[]{\includegraphics[width=0.32\textwidth]{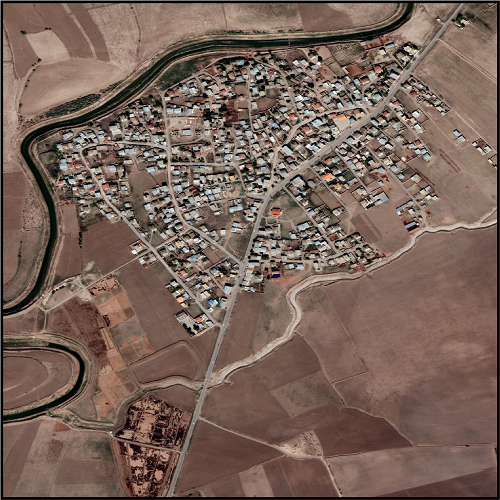}
    \label{iranb}}
    \hfil
    \subfigure[]{\includegraphics[width=0.32\textwidth]{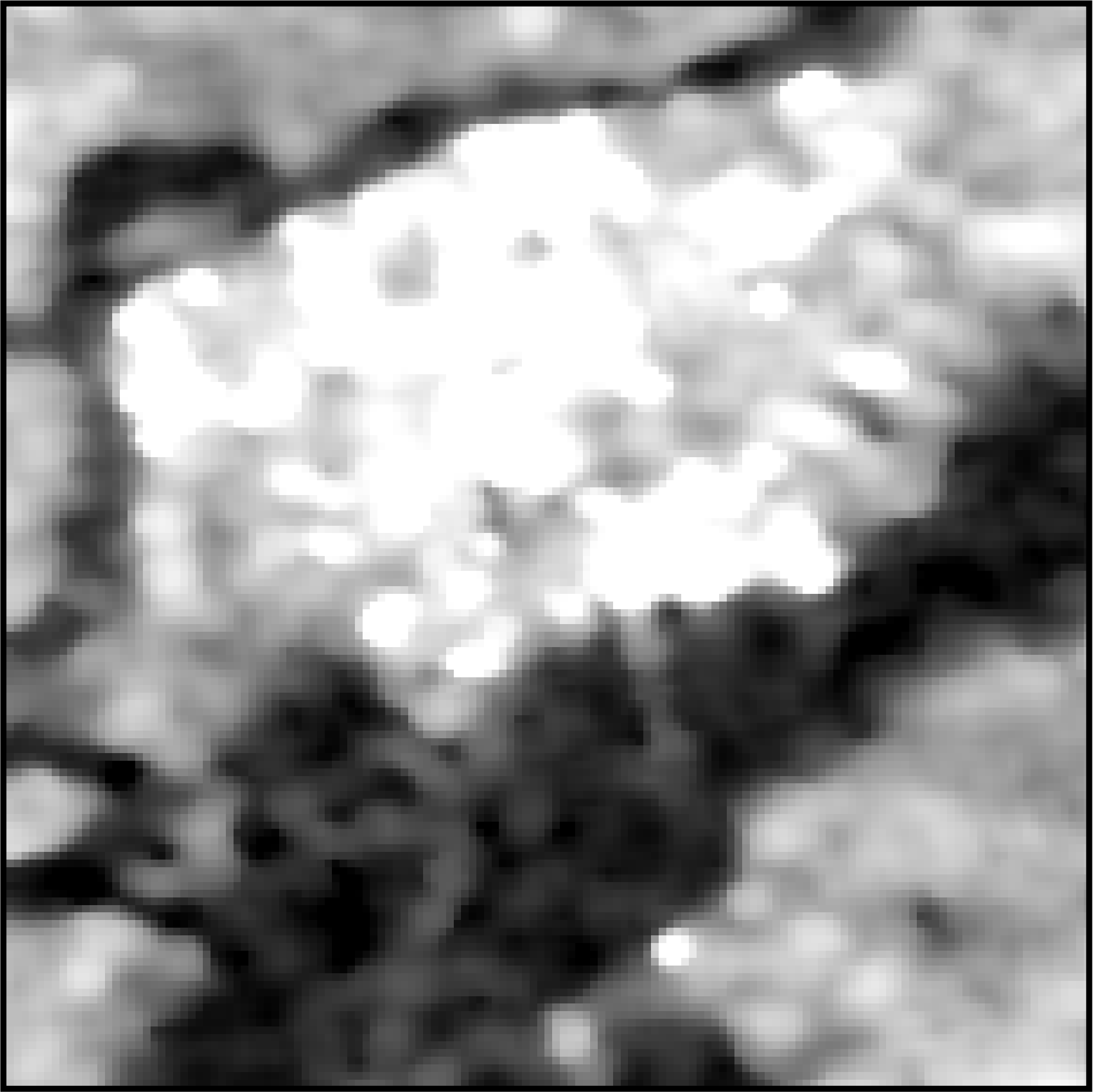}
    \label{iranc}}
    \hfil
    \subfigure[]{\includegraphics[width=0.32\textwidth]{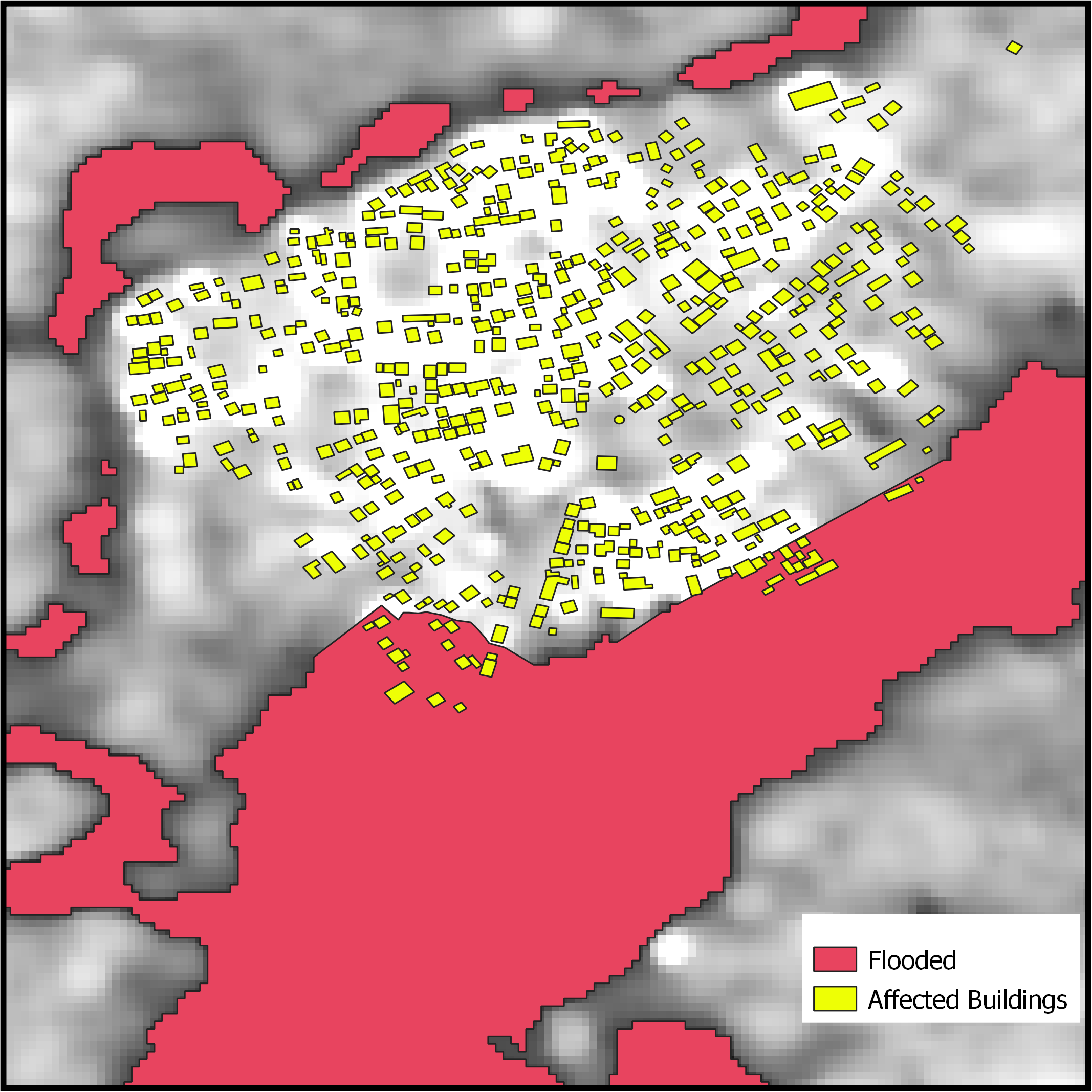}
    \label{irand}}

    \caption{Flood inundation mapping of the Iran region. \subref{irana} The Iran region has a total area of 261.638 sq km, and the flooding occurred on January 13, 2020. \subref{iranb}-\subref{irand} are zoomed-in views of the pre-flood optical image, post-flood SAR image, and flood mapping of yellow box in \subref{irana}, respectively.}
    \label{testingmap2}
\end{figure*}


\subsection{Large-scale Flood Inundation Mapping}
\label{flood mapping}
After the model performance verification, we adopt the proposed DAM-Net for large-scale flood inundation mapping. Sentinel-1 SAR images collected from Nebraska and Iran are used in this experiment. Specifically, the Nebraska region was imaged on 18 March 2019, with 10 m spatial resolution covering 613.581 square kilometers. The flood event was caused by hurricane and rapid snowmelt, resulting in over \$1.3 billion in damages. Figure~\ref{testingmap1} displays the flood inundation map and two zoomed-in views as typical. We also provide the optical image before the flood event and the SAR image during the flood event as references, for the intuitive mapping effect. Figure~\ref{testingmap2} shows the flood inundation map of southern Iran. The region was imaged on 13 January 2020 covering 261.638 square kilometers, which suffered flooding caused by severe rainfall and cyclones and faced flood damages of 196,140 affected people. These results further demonstrate that the proposed DAM-Net is able to boost the flood detection task. Expect flooded areas, the common land-cover category, such as buildings and permanent water, can be marked in the flood inundation map, since the \textit{S1GFloods} dataset additionally provides land-cover information. This contributes to post-flood evacuation, protection planning, and disaster assessment.

\begin{table*}[!ht]
\centering
\caption{Ablation study of the proposed DAM-Net on the \textit{S1GFloods} dataset. \textit{OEL} and \textit{CA-CE} denote our overlapping embedding layer and the combination of CTCA and TACE modules, while \textit{ST} semantic token. We highlight the performance gain and best result in \textcolor{red}{red}, \textbf{bold}, respectively.}
\renewcommand\arraystretch{1.2}
\begin{tabular*}{\tblwidth}{@{}LCCCCCC @{}}
\hline
\multirow{2}{*}{Models}     & \multicolumn{3}{c}{Configuration} & \multirow{2}{*}{OA (\%)} & \multirow{2}{*}{F1 (\%)} & \multirow{2}{*}{IoU (\%)} \\
\cline{2-4}
         & Overlap. Embed.      & CTCA+TACE & Sem. Token &                      &                      &                      \\
\hline 
Baseline          &            &            &            & 95.6 & 90.2 & 86.3                     \\
+\textit{OEL}     & \checkmark &            &            & 96.9 \scriptsize{\textcolor{red}{(+1.3)}} & 92.3 \scriptsize{\textcolor{red}{(+2.1)}} & 88.1 \scriptsize{\textcolor{red}{(+1.8)}} \\
+\textit{CA-CE}   &            & \checkmark &            & 98.2 \scriptsize{\textcolor{red}{(+2.6)}} & 94.9 \scriptsize{\textcolor{red}{(+4.7)}} & 91.8 \scriptsize{\textcolor{red}{(+5.5)}} \\
+\textit{ST}      &            &            & \checkmark & 98.1 \scriptsize{\textcolor{red}{(+2.5)}} & 94.6 \scriptsize{\textcolor{red}{(+4.4)}} & 91.7 \scriptsize{\textcolor{red}{(+5.4)}} \\
DAM-Net           & \checkmark & \checkmark & \checkmark & \textbf{98.8} \scriptsize{\textcolor{red}{(+3.2)}} & \textbf{95.3} \scriptsize{\textcolor{red}{(+5.1)}}& \textbf{92.6} \scriptsize{\textcolor{red}{(+6.3)}} \\     
\hline
\end{tabular*}
\label{tab-ablation}
\end{table*}

\begin{figure*}[ht]
     \centering
     \includegraphics[width=0.798\textwidth]{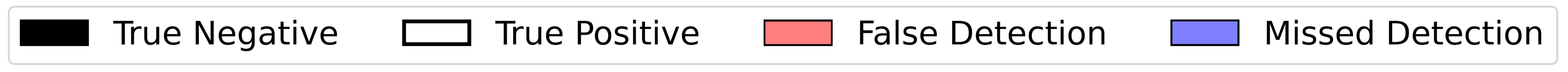}
     \\
     \includegraphics[width=0.998\textwidth]{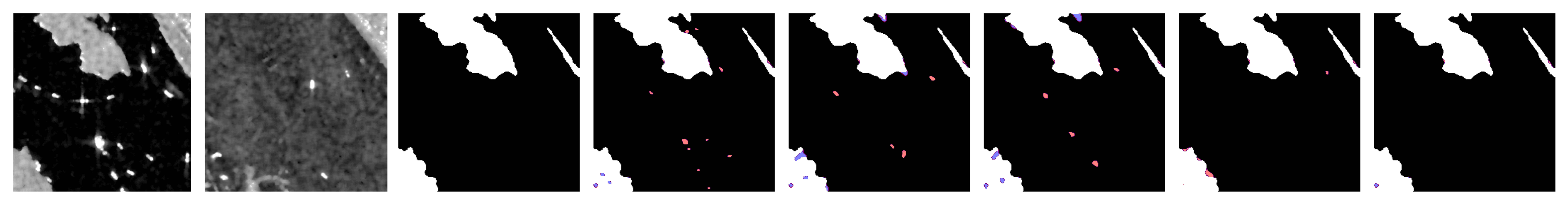}
     \\
     \includegraphics[width=0.998\textwidth]{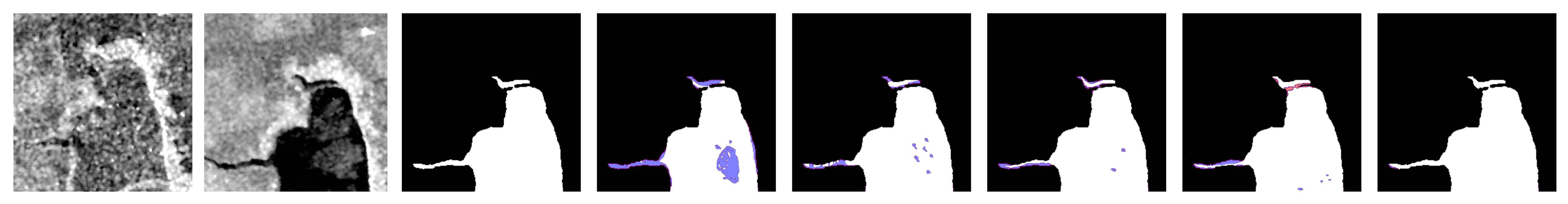}
     \\
     \includegraphics[width=0.998\textwidth]{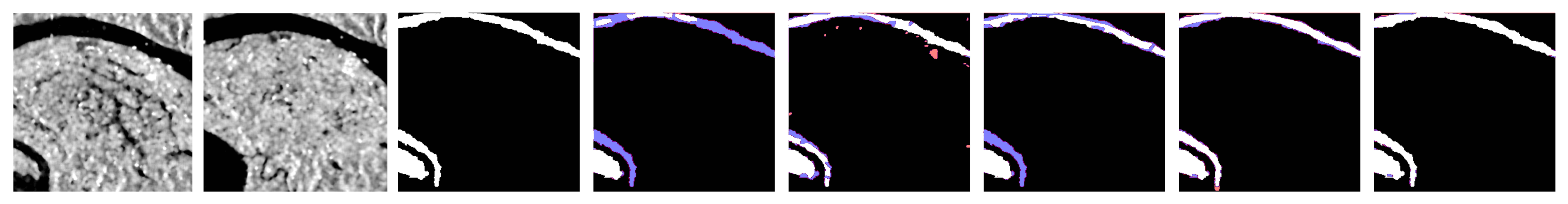}
     \\
     \includegraphics[width=0.998\textwidth]{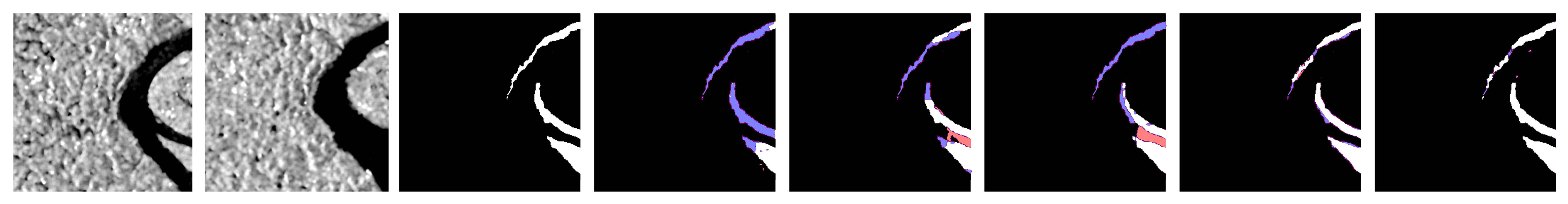}
     \\
     \includegraphics[width=0.998\textwidth]{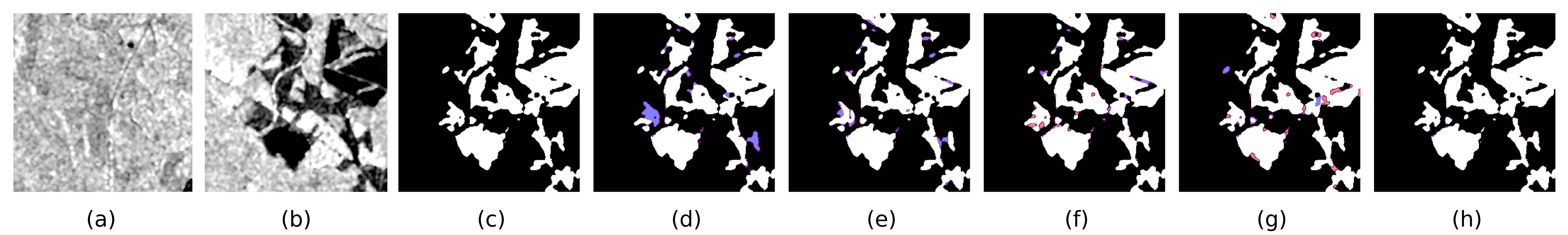}  
     \caption{Visualization results of the ablation experiment. (a) the pre-flood image, (b) the post-flood image, (c) the ground truth, (d)-(h) are detection results of Baseline, Baseline+\textit{OEL}, Baseline+\textit{CA-CE}, Baseline+\textit{ST}, and DAM-Net, respectively.}
     \label{fig:ablation}
\end{figure*}

\subsection{Ablation Study on Module Effect}
\label{Ablation Study}
We conduct ablation studies to verify the effectiveness of key design, including the overlapping embedding layer, the combination of cross-temporal change attention and temporal-aware change enhancement, and the semantic token, as shown in Figure~\ref{fig:ablation} and Table~\ref{tab-ablation}. The baseline contains the original ViTAEv2 to extract multi-scale features, a feature fusion module that implements absolute difference and concatenation operations to obtain change features, and the prediction head to achieve the final flood map. We first add to the baseline our overlapping embedding layer: this  preserves local continuity (see the third image pair in Figure~\ref{fig:ablation} and provides a boost on all performance metrics (1.3\% OA, 2.1\% F1, and 1.8\% IoU). Second, we add CTCA and TACE modules to the baseline. The performance improvement is remarkable, especially on IoU (5.5\%). This confirms the importance of identifying change features using global contextual information (CTCA module) and accounting for the information loss issue (TACE module). Finally, we introduce the semantic token. This component increases the ability to distinguish water-body changes from various changes. We can see in the first image pair of Figure~\ref{fig:ablation} that the changes caused by temporary objects (e.g., ships) are effectively suppressed. Consequently, DAM-Net with all modules outperforms the baseline by a large margin, achieving 98.8\% OA, 95.3\% F1, and 92.6\% IoU.

\section{Conclusions}
\label{Conclusions}
This paper addresses the flood detection problem using high-resolution Sentinel-1 SAR image pairs captured during global flood events over the past eight years. A new DAM-Net model is proposed to tackle the challenging task of flood detection in complex backgrounds. The temporal-differential fusion module in DAM-Net provides indispensable information for exploring the relationship between semantic tokens and change features, greatly enhancing differential image features and producing more accurate flood maps. Additionally, a new dataset called \textit{S1GFloods} is created to train and evaluate flood detection models for Sentinel-1 SAR image pairs. \textit{S1GFloods} contains a wider range of diverse flood events involving different causes, such as heavy rains, overflowing rivers, broken dams, tropical storms, and hurricanes. It also includes richer inundation scenes, encompassing wetlands, riverine areas, mountainous regions, urban and rural areas, and vegetation. This dataset is beneficial for developing flood detection methods with excellent performance and good generalizability. The results demonstrate that CNN-based flood detection methods performed relatively poorly on the newly created \textit{S1GFloods} dataset compared to effective vision transformers, with an IoU score that was 7.8\% lower. The experiments further validate the effectiveness and superiority of our proposed method across all evaluation metrics when compared with previous methods. Based on the findings, the authors recommend improving the quality of the images used in the \textit{S1GFloods} dataset, utilizing information from another SAR sensor more effectively, and acquiring images from drones or cloud-free optical remote-sensing sources to increase the accuracy of water body interpretation.

\section*{Acknowledgment}
Tamer~Saleh is supported by the China Scholarship Council (CSC). 


\bibliographystyle{cas-model2-names}
\bibliography{cas-refs}

\end{document}